\documentclass{article}
\usepackage{ijcai26}

\usepackage{times}
\usepackage{soul}
\usepackage{url}
\usepackage[hidelinks]{hyperref}
\usepackage[utf8]{inputenc}
\usepackage[small]{caption}
\usepackage{graphicx}
\usepackage{amsmath}
\usepackage{amsthm}
\usepackage{booktabs}
\usepackage{algorithmicx}   
\usepackage{algpseudocode}  
\usepackage{algorithm}
\usepackage[switch]{lineno}
\usepackage{siunitx}      
\usepackage{multirow}         
\usepackage{graphicx}
\usepackage{bm}
\usepackage{amsfonts}
\usepackage{multicol}
\usepackage{bbm}
\usepackage{adjustbox}
\DeclareMathOperator*{\argmin}{argmin}
\usepackage{dirtytalk}
\title{Vector-Quantized Discrete Latent Factors Meet Financial Priors: Dynamic Cross-Sectional Stock Ranking Prediction for Portfolio Construction}

\author{
Namhyoung Kim$^{1,2}$ \and
Jae Wook Song$^{2}$
\affiliations
$^1$RiskX, $^2$Hanyang University\\
\emails
namhyoung.kim@riskx.tech, jwsong@hanyang.ac.kr
}

\begin{document}

\maketitle

\begin{abstract}
Predicting cross-sectional stock returns is challenging due to low signal-to-noise ratios and evolving market regimes. Classical factor models offer interpretability but limited flexibility, while deep learning models achieve strong performance yet often underutilize financial priors. We address this gap with \textbf{PRISM-VQ} (\underline{PR}ior-\underline{I}nformed \underline{S}tock \underline{M}odel with \underline{V}ector \underline{Q}uantization), a dynamic factor framework that integrates expert prior factors, vector-quantized discrete latent factors learned from cross-sectional structure, and a structure-conditioned Mixture-of-Experts to generate time-varying factor loadings. Vector quantization acts as an information bottleneck that suppresses noise while capturing robust market structure, with discrete codes serving both as latent factors and as routing signals for temporal expert specialization. Experiments on CSI~300 and S\&P~500 show consistent improvements in cross-sectional return prediction and portfolio performance over strong baselines while preserving interpretability. Our code is available at \texttt{\url{https://github.com/finxlab/PRISM-VQ}}.
\end{abstract}

\section{Introduction}
Predicting cross-sectional stock returns remains a fundamental challenge at the intersection of finance and machine learning due to persistently low signal-to-noise ratios (SNRs), complex cross-asset dependencies, and frequent regime shifts~\cite{israel2020can,zhang2025major}. Classical linear factor models~\cite{sharpe1964capital,fama1992cross,fama1993common} provide strong interpretability through economically grounded factors and loadings, but their limited expressiveness restricts their ability to capture nonlinear interactions and time-varying dynamics. This limitation has contributed to the proliferation of the so-called ``factor zoo''~\cite{cochrane2011presidential}, highlighting the difficulty of robust factor discovery.

Recent neural approaches to cross-sectional return prediction largely fall into two categories. \textbf{Factor-based methods} integrate neural networks within factor model structures. Autoencoder-based approaches~\cite{gu2021autoencoder,duan2022factorvae,kim2025factorvqvae} learn latent factors via reconstruction objectives, but their reliance on continuous latent representations often provides insufficient regularization under low-SNR conditions and struggles with temporal non-stationarity. In contrast, \textbf{architecture-centric deep learning methods} employ modern sequence models such as Transformers~\cite{yoo2021accurate,li2024master,cao2024matcc} to model spatio-temporal dependencies directly from data. Despite their expressive power, these models typically lack explicit structural priors, which reduces robustness under distribution shifts and limits interpretability.

\begin{figure}[t] 
    \centering
    \includegraphics[width=0.95 \linewidth]{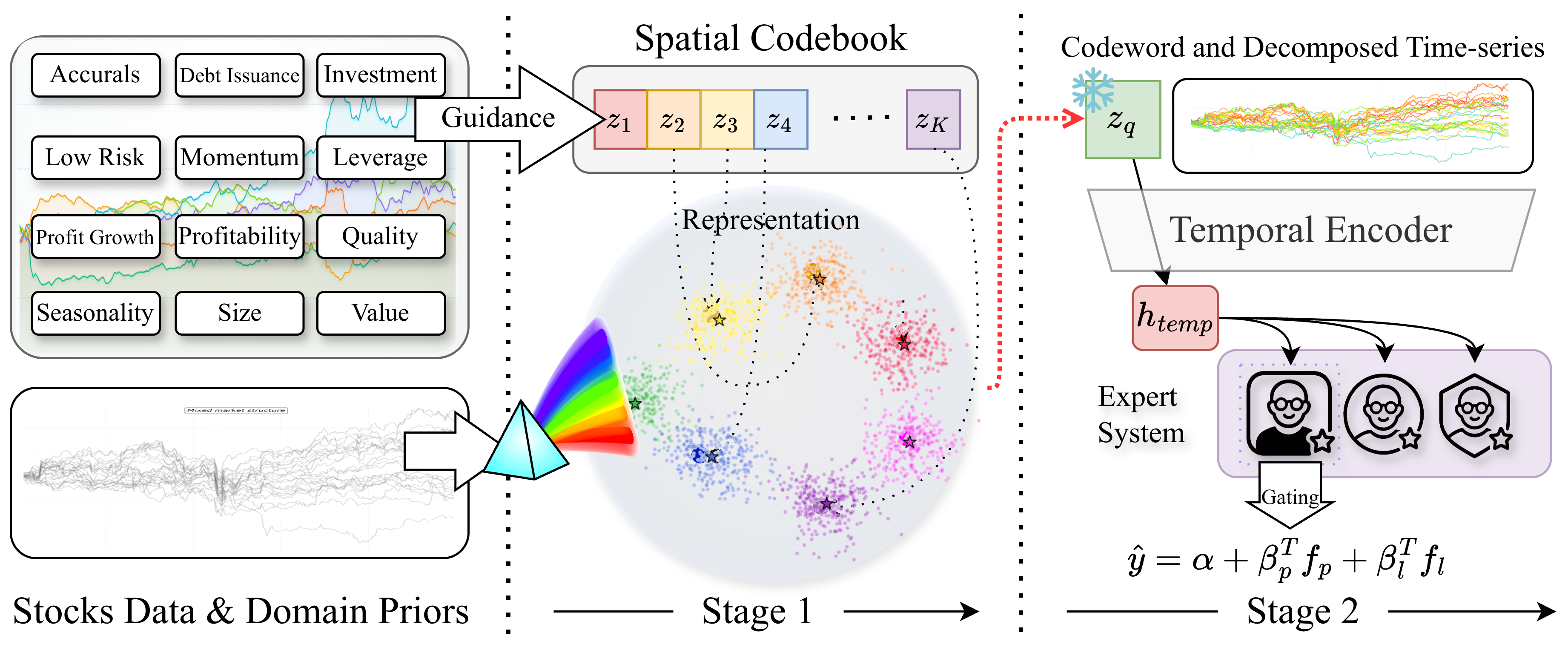} 
    \caption{\textbf{Overview of the PRISM-VQ framework.} Like a prism separating light, PRISM-VQ decomposes cross-sectional market structure using discrete vector quantization.}
    \label{fig:overall}
\end{figure}

Across prior work, three recurring limitations emerge: (1)~\emph{insufficient regularization} of continuous latent representations in noisy environments; (2)~\emph{structure-agnostic temporal modeling} that fails to condition on evolving cross-sectional structure; and (3)~\emph{underutilization of domain priors}, where established financial knowledge is weakly exploited despite its stabilizing potential.

To address these limitations, we propose \textbf{PRISM-VQ}
(\underline{PR}ior-\underline{I}nformed \underline{S}tock \underline{M}odel with \underline{V}ector \underline{Q}uantization), a unified framework for cross-sectional stock return prediction that integrates discrete representation learning, structure-conditioned temporal modeling, and domain-informed priors within a factor-based interpretation (Figure~\ref{fig:overall}). PRISM-VQ consists of three components. First, it learns \textbf{vector-quantized latent factors} through a joint vector quantization and contrastive learning objective~\cite{chen2020simple}, where the learned codebook~\cite{van2017neural} serves as an information bottleneck that suppresses noise while preserving cross-sectional structure. Second, it introduces a \textbf{structure-conditioned Mixture-of-Experts} (MoE)~\cite{shazeer2017outrageously} built on Transformer architectures, where discrete codes act as gating signals to generate dynamic factor loadings. Third, it incorporates \textbf{domain-specific prior factors}~\cite{jensen2023there} as stabilizing anchors that maintain interpretability and improve robustness under distribution shifts.

We summarize our contributions as follows:
\begin{itemize}
    \item We propose PRISM-VQ, a factor-based framework that combines financial prior factors, vector-quantized discrete latent factors, and structure-conditioned temporal modeling for cross-sectional stock rank prediction.
    \item We demonstrate that vector quantization provides an effective inductive bias for cross-sectional representation learning, consistently outperforming continuous latent alternatives in noisy financial settings.
    \item We introduce a discrete-code-gated MoE mechanism that enables structure-conditioned dynamic factor loading generation while preserving a clear factor interpretation and supporting scalable conditional computation.
    \item We conduct experiments on CSI~300 and S\&P~500, where PRISM-VQ achieves the state-of-the-art performance, including RankIC of 0.0646 and 0.0141, and Sharpe ratios of 1.57 and 0.67, respectively.
\end{itemize}

\section{Related Works}
\label{sec:related}

\subsection{Machine Learning-Based Factor Models}
Factor models have long been central to asset pricing, from the CAPM~\cite{sharpe1964capital} to the Fama--French models~\cite{fama1992cross}.
Although interpretable, their linear structure limits expressiveness in modern markets. Autoencoder asset pricing~\cite{gu2021autoencoder} introduced neural networks to learn nonlinear factor structures, followed by extensions such as FactorVAE~\cite{duan2022factorvae} and FactorVQVAE~\cite{kim2025factorvqvae}, which employs vector quantization~\cite{van2017neural} for discrete latent factors.
Later work incorporated additional structure, including graph-based industry modeling~\cite{duan2025factorgcl} and regime-aware dynamics~\cite{xiang2024rsap}.
Despite these advances, most methods remain autoencoder-based, limiting temporal modeling capacity and underutilizing financial priors. PRISM-VQ addresses these limitations by combining discrete factor learning with Transformer-based temporal modeling and explicit domain-informed priors.

\subsection{Deep Learning for Stock Prediction}
Another line of work adopts end-to-end deep learning without explicit factor structures.
Transformer-based models have been particularly influential.
DTML~\cite{yoo2021accurate} modeled inter-stock dependencies with attention, MASTER~\cite{li2024master} alternated intra- and inter-stock attention, and MATCC~\cite{cao2024matcc} incorporated trend decomposition.
While effective, these models typically lack interpretable factor representations, weakly encode domain knowledge, and show limited robustness under regime shifts.
More recently, explicit relational modeling has been explored through temporal-relational graphs and hypergraph structures~\cite{chen2024automatic,song2025multi,alaygut2025hypergraph}. 
These methods rely on predefined or extracted relational topologies, whereas PRISM-VQ learns implicit cross-sectional structure through a discrete codebook without requiring relational preprocessing, enabling transferability across markets.
PRISM-VQ retains the expressive power of Transformers while preserving factor interpretability by using them to generate dynamic factor loadings conditioned on learned cross-sectional structure.

\subsection{Representation Learning for Finance}
The low SNRs of financial data place strong demands on representation learning.
Contrastive learning has been explored to learn robust stock embeddings~\cite{du2024explainable,hwang2023simstock}.
Vector quantization offers a complementary approach by introducing a discrete information bottleneck that suppresses noise and promotes stable pattern reuse.
Recent work has applied vector quantization to identify persistent market structure; for example, STORM~\cite{zhao2024storm} employs dual VQ-VAEs to capture spatio-temporal patterns, while FactorVQVAE~\cite{kim2025factorvqvae} introduces discrete latent factors via a single-stage VQ objective. PRISM-VQ differs from both by combining contrastive-regularized VQ with structure-conditioned MoE routing and explicit financial priors within a unified two-stage framework.
To our knowledge, PRISM-VQ is the first framework to integrate vector-quantized discrete factors, contrastive representation learning, expert prior factors, and discrete-code-conditioned temporal modeling within a unified factor-based modeling framework.


\section{Problem Formulation}
\label{sec:preliminaries}

We study cross-sectional stock return prediction. At each decision time $t$, the objective is to predict returns for $N_t$ stocks over the next $\Delta$ days for ranking-based portfolio construction. Unless otherwise specified, all quantities are measured at time $t$, and we omit the explicit time index in symbols such as $x_i$, $X$, $f_p$, $y_i$, and $\hat y_i$.

\subsection{Data and Factor Model Framework}
Each stock $i$ is associated with temporal features $x_i \in \mathbb{R}^{T\times C}$, where $T$ is the lookback window and $C$ is the feature dimension.
Stacking all stocks yields the cross-sectional tensor
$X=\{x_i\}_{i=1}^{N_t}\in\mathbb{R}^{N_t\times T\times C}$.
We also observe $P$ expert prior factors $f_p\in\mathbb{R}^{P}$, such as value, size, and momentum. The prediction target $y_i$ corresponds to the return from $t+1$ to $t+\Delta$.

Returns are modeled using a dynamic factor structure:
\begin{equation}
y_i = \alpha_i + \beta_{p,i}^T f_p + \beta_{l,i}^T f_{l,i} + \epsilon_i,
\end{equation} where $\alpha_i$ is the intercept, $\beta_{p,i}$ and $\beta_{l,i}$ are loadings on prior and learned factors, $f_{l,i}$ denotes learned latent factors, and $\epsilon_i$ is idiosyncratic noise. Unlike classical factor models with static linear loadings, both loadings and latent factors are generated by nonlinear mappings:
\begin{align}
(\alpha_i, \beta_{p,i}, \beta_{l,i}) &= \Phi_{\text{loading}}(x_i, f_p; \theta), \\
f_{l,i} &= \Phi_{\text{factor}}(x_i, f_p; \theta),
\end{align} where $\Phi_{\text{loading}}$ and $\Phi_{\text{factor}}$ are deep neural networks with parameters $\theta$. 

\subsection{Modeling Challenges}
Cross-sectional stock return prediction faces several challenges: (i) extremely low SNRs, (ii) complex and evolving cross-sectional dependencies, (iii) non-stationary market regimes requiring adaptive factor loadings, and (iv) the need for interpretability consistent with financial theory. PRISM-VQ addresses these challenges through a two-stage learning paradigm, where the spatial learning stage discovers discrete latent factors via vector quantization and the temporal learning stage generates dynamic factor loadings using structure-conditioned expert networks.

\section{Proposed Model}
\label{sec:method}

\subsection{Overview: Two-Stage Learning}
\begin{figure*}[!t]
    \centering
    \includegraphics[width=\textwidth]{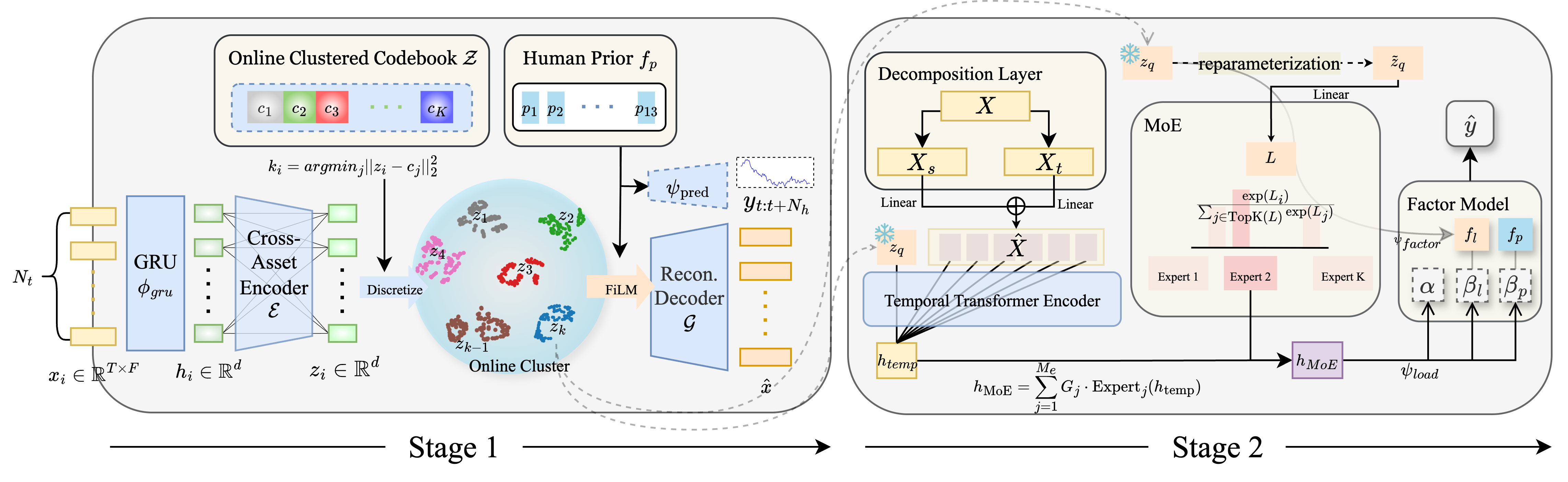}
    \caption{\textbf{Architecture of PRISM-VQ.} The spatial learning stage (left) learns discrete stock representations via vector quantization over cross-sectional features. The temporal learning stage (right) uses the resulting discrete codes to gate expert networks and generate dynamic factor loadings that combine expert prior factors with learned latent factors for return prediction.}
    \label{fig:fpq-detail}
\end{figure*}

PRISM-VQ decouples \emph{cross-sectional structure discovery} from \emph{time-varying factor loading generation} through a two-stage framework. Stage~1 (\textbf{Spatial Learning}) learns a discrete codebook and stock-level assignments via vector quantization, mapping codes to latent factor values. Stage~2 (\textbf{Temporal Learning}) generates dynamic factor loadings using a Transformer-based MoE, with expert routing conditioned on the discrete codes. After Stage~1, the codebook is fixed, while stock assignments may vary over time (Figure~\ref{fig:fpq-detail}).

\paragraph{Stage 1: Code assignment.}
At time $t$, Stage~1 models discrete code assignments conditioned on cross-sectional features and prior factors:
\small
\begin{align} 
p(\{z_{q,i}\}_{i=1}^{N_t} \mid \{x_i\}_{i=1}^{N_t}, f_p)
&= \prod_{i=1}^{N_t} q\!\left(z_{q,i} \mid \{x_j\}_{j=1}^{N_t}, f_p\right).
\label{eq:stage1_assign}
\end{align} 
\normalsize
Each discrete code $z_{q,i}\in\mathbb{R}^{d_s}$ is then mapped to a latent factor value
\begin{align}
f_{l,i} = \psi_{\text{factor}}(z_{q,i}) \in \mathbb{R}^{d_s}.
\label{eq:factor_map}
\end{align}

\paragraph{Stage 2: Code-conditioned loading generation.}
Let $\boldsymbol{\beta}_i = (\alpha_i, \beta_{p,i}, \beta_{l,i})$ denote the full set of factor loadings. Conditioned on the discrete code $z_{q,i}$, Stage~2 generates dynamic loadings via a code-conditioned MoE:
\begin{align}
p(\boldsymbol{\beta}_i \mid h_{\text{temp},i}, z_{q,i})
= \sum_{j=1}^{M_e} p(e_j \mid z_{q,i})\,p_{\theta_j}(\boldsymbol{\beta}_i \mid h_{\text{temp},i}),
\label{eq:stage2_mixture}
\end{align} thereby conditioning temporal modeling on learned cross-sectional structure while preserving a factor interpretation.

\subsection{Spatial Learning: Discrete Cross-Sectional Factor Discovery}
\label{sec:spatial}

\subsubsection{Cross-Sectional Feature Extraction}
Each stock $i$ is associated with features $x_i \in \mathbb{R}^{T\times C}$. We first encode temporal dynamics using a GRU encoder $\phi_{\text{gru}}$:
\begin{align}
h_i = \phi_{\text{gru}}(x_i) \in \mathbb{R}^{d_s}.
\label{eq:gru_enc}
\end{align}

To capture cross-sectional interactions at time $t$, we stack all embeddings into
$H=[h_1;\ldots;h_{N_t}] \in \mathbb{R}^{N_t\times d_s}$ and apply a cross-asset Transformer encoder $\mathcal{E}$:
\begin{align}
z = \mathcal{E}(H) \in \mathbb{R}^{N_t\times d_s}, \quad z_i \in \mathbb{R}^{d_s}.
\label{eq:cross_asset}
\end{align}

Through self-attention, each $z_i$ aggregates information from related stocks, forming a representation suitable for discrete clustering.

\subsubsection{Vector Quantization for Discrete Factor Learning}
We discretize the continuous embeddings using a learnable codebook
$\mathcal{Z}=\{\bm{c}_k\}_{k=1}^{K}\subset\mathbb{R}^{d_s}$.
Each embedding is assigned to its nearest codeword:
\begin{align}
k_i = \argmin_{k\in\{1,\ldots,K\}} \bigl\lVert z_i-\bm{c}_k \bigr\rVert_2^2,
\quad
z_{q,i} = \bm{c}_{k_i}.
\label{eq:vq_assign}
\end{align}

This discretization introduces a \emph{finite prototype space} that acts as a strong inductive bias under low SNRs. It (i) suppresses idiosyncratic noise by snapping embeddings to recurring cross-sectional prototypes, (ii) limits effective model capacity through a bounded set of codewords, and (iii) encourages sample sharing among stocks assigned to the identical code, yielding stable and interpretable clusters. Since codewords are reused over time, the learned codes naturally separate persistent cross-sectional structures from time-varying factor sensitivities in the two-stage design.


To enable end-to-end training despite the non-differentiable assignment, we adopt the straight-through estimator~\cite{van2017neural} and optimize
\begin{align}
\mathcal{L}_{\text{VQ}}
= \|\text{sg}(z)-z_q\|_2^2
+ \lambda_{\text{commit}}\|z-\text{sg}(z_q)\|_2^2,
\label{eq:vq_loss}
\end{align} where $\text{sg}(\cdot)$ denotes the stop-gradient operator.

\subsubsection{Contrastive Learning for Semantic Clustering}
Nearest-neighbor quantization alone may yield arbitrary partitions. We therefore introduce a codebook contrastive objective~\cite{chen2020simple}:
\begin{align}
\mathcal{L}_{\text{contra}}
= -\log
\frac{\exp(\text{sim}(z_i,\bm{c}_{k_i})/\tau)}
{\sum_{k=1}^{K}\exp(\text{sim}(z_i,\bm{c}_{k})/\tau)},
\label{eq:contra}
\end{align} where $\tau$ is a temperature parameter and $\text{sim}(u,v)=-\|u-v\|_2$. This objective contrasts each embedding against all codeword prototypes, explicitly enlarging the margin between the assigned codeword and competing codewords. As a result, clustering is guided by cross-sectional semantics rather than purely geometric proximity.

\subsubsection{Reconstruction and Auxiliary Prediction}
Pure clustering can produce stable but weakly predictive representations, while pure prediction may fail to extract reusable structure. We therefore introduce two auxiliary tasks to regularize the discrete codes, $z_{q,i}$.

\paragraph{Reconstruction.}
A decoder $\mathcal{G}$ reconstructs the input features from the quantized code, conditioned on prior factors using FiLM (Feature-wise Linear Modulation)~\cite{perez2018film}:
\begin{align}
\hat{x}_i &= \mathcal{G}(z_{q,i}, f_p), \nonumber\\
\mathcal{L}_{\text{recon}}
&= \frac{1}{N_t}\sum_{i=1}^{N_t}\|x_i-\hat{x}_i\|_2^2.
\label{eq:recon}
\end{align}
This conditioning encourages the learned codes to remain consistent with the expert prior information.

\paragraph{Multi-horizon prediction.}
To retain forward-looking signals, we further regularize the codes through multi-horizon return prediction:
\begin{align}
[& \hat{y}_{i,1},\ldots,\hat{y}_{i,N_h}] = \psi_{\text{pred}}(z_{q,i}, f_p), \nonumber\\
& \mathcal{L}_{\text{pred}} = \frac{1}{N_tN_h}\sum_{i=1}^{N_t}\sum_{h=1}^{N_h}(y_{i,h}-\hat{y}_{i,h})^2.
\label{eq:pred}
\end{align}

\subsubsection{Spatial Learning Objective}
Then, the following multi-task objective is optimized:
\begin{align}
\mathcal{L}_{\text{spatial}}
= \mathcal{L}_{\text{recon}} + \mathcal{L}_{\text{VQ}}
+ \lambda_{\text{contra}}\mathcal{L}_{\text{contra}}
+ \lambda_{\text{pred}}\mathcal{L}_{\text{pred}}.
\label{eq:spatial_obj}
\end{align}
Stage~1 outputs the discrete codes $z_{q,i}$ and the corresponding latent factor values $f_{l,i}$ via Eq.~\eqref{eq:factor_map}, which are subsequently used in Stage~2.

\subsection{Temporal Learning: Structure-Conditioned Dynamic Factor Loadings}
\label{sec:temporal}

\subsubsection{Temporal Context Encoding with a Structure Token}
Given $z_{q,i}$ from Stage~1, we encode temporal context to generate dynamic factor loadings. Let $z_{q,i}\in\mathbb{R}^{d_s}$ and let the temporal encoder operate in $\mathbb{R}^{d_t}$. We optionally decompose each input sequence into trend and residual components and project them to dimension $d_t$:
\begin{align}
x_{i,\text{trend}} &= \text{AvgPool}_{w}(\text{Pad}(x_i)), \nonumber\\
x_{i,\text{seasonal}} &= x_i - x_{i,\text{trend}}, \nonumber\\
\tilde{x}_i &= \phi_{\text{seasonal}}(x_{i,\text{seasonal}}) + \phi_{\text{trend}}(x_{i,\text{trend}}).
\label{eq:decomp}
\end{align}
To inject cross-sectional structure, we prepend $z_{q,i}$ as a structure token (projected to $d_t$ if $d_s\neq d_t$) and apply a temporal Transformer encoder $\mathcal{T}$ with RoPE~\cite{su2024roformer}:
\begin{align}
h_{\text{temp},i} = \mathcal{T}([z_{q,i};\tilde{x}_i]) \in \mathbb{R}^{d_t}.
\label{eq:temp_enc}
\end{align}

\subsubsection{Structure-Conditioned MoE for Dynamic Loadings and Return Prediction}
Dynamic factor loadings are generated via a structure-conditioned MoE, where $z_{q,i}$ determines expert routing.

\paragraph{Stochastic gating mechanism.}
We compute the expert-wise mean and scale parameters from the discrete code:
\begin{align}
\mu_i &= \phi_\mu(z_{q,i}) \in \mathbb{R}^{M_e}, \nonumber\\
\sigma_i &= \mathrm{Softplus}(\phi_\sigma(z_{q,i})) \in \mathbb{R}^{M_e},
\label{eq:gate_mu_sigma}
\end{align} and sample pre-activations via reparameterization:
\begin{align}
\tilde{g}_i = \mu_i + \epsilon_i \odot \sigma_i, \quad \epsilon_i \sim \mathcal{N}(0, I_{M_e}).
\label{eq:gate_reparam}
\end{align}
We obtain logits $L_i = \psi_g(\tilde{g}_i) \in \mathbb{R}^{M_e}$, select the top-$k$ experts $\mathcal{K}_i = \mathrm{Top}_k(L_i)$, and apply a local softmax to define the gating weights:
\begin{align}
G_{i,j} =
\begin{cases}
\displaystyle\frac{\exp(L_{i,j})}{\sum_{\ell\in\mathcal{K}_i}\exp(L_{i,\ell})} & \text{if } j\in\mathcal{K}_i,\\[6pt]
0 & \text{otherwise.}
\end{cases}
\label{eq:topk_softmax}
\end{align}
By construction, $\sum_j G_{i,j}=1$ and $\|G_i\|_0=k$.

\paragraph{Expert aggregation and loading projection.}
Each expert $\xi_j:\mathbb{R}^{d_t}\rightarrow\mathbb{R}^{d_t}$ is implemented as a lightweight MLP:
\begin{align}
\xi_j(h) = W_j^{(2)}\,\mathrm{GELU}(W_j^{(1)}h+b_j^{(1)}) + b_j^{(2)}.
\label{eq:expert_mlp}
\end{align}

Expert outputs are aggregated and projected to dynamic factor loadings:
\begin{align}
h_{\text{MoE},i} &= \sum_{j=1}^{M_e} G_{i,j}\,\xi_j(h_{\text{temp},i}), \nonumber\\
(\alpha_i,\beta_{p,i},\beta_{l,i}) &= \psi_{\text{load}}(h_{\text{MoE},i}).
\label{eq:moe_and_load}
\end{align}

\paragraph{Return prediction.}
We compute the learned factor values from the discrete codes and predict returns:
\begin{align}
f_{l,i} &= \psi_{\text{factor}}(z_{q,i}), \nonumber\\
\hat{y}_i &= \alpha_i + \beta_{p,i}^{\top} f_p + \beta_{l,i}^{\top} f_{l,i},
\label{eq:final_pred}
\end{align} where $\alpha_i\in\mathbb{R}$, $\beta_{p,i}\in\mathbb{R}^{P}$, $\beta_{l,i}\in\mathbb{R}^{d_s}$, and $f_p\in\mathbb{R}^{P}$.

\subsubsection{Training Objective with Load-Balancing Regularization}
The temporal learning stage minimizes the mean squared error with a standard load-balancing regularizer~\cite{shazeer2017outrageously} to prevent expert collapse:
\begin{equation}
\begin{aligned}
f_j &= \frac{1}{N_t}\sum_{i=1}^{N_t}\mathbb{I}[j\in\mathcal{K}_i], \qquad
P_j = \frac{1}{N_t}\sum_{i=1}^{N_t} G_{i,j},\\
\mathcal{L}_{\text{temporal}}
&= \frac{1}{N_t}\sum_{i=1}^{N_t}(\hat{y}_i-y_i)^2
+ \lambda_{\text{balance}}\, M_e\sum_{j=1}^{M_e} f_j P_j .
\end{aligned}
\label{eq:temporal_obj}
\end{equation}

\subsection{Two-Stage Training Procedure}

\paragraph{Stage 1: Spatial learning.}
We optimize $\mathcal{L}_{\text{spatial}}$ in Eq.~\eqref{eq:spatial_obj} to learn the temporal encoder $\phi_{\text{gru}}$, cross-asset encoder $\mathcal{E}$, codebook $\mathcal{Z}$, and auxiliary modules $\mathcal{G}$ and $\psi_{\text{pred}}$.
This stage produces discrete codes $z_{q,i}$ and latent factor values $f_{l,i}$.

\paragraph{Stage 2: Temporal learning.}
We fix the learned codebook $\mathcal{Z}$ and optimize $\mathcal{L}_{\text{temporal}}$ in Eq.~\eqref{eq:temporal_obj} to train the temporal Transformer $\mathcal{T}$, MoE gating and experts, and the loading projection $\psi_{\text{load}}$.
Discrete codes from Stage~1 act as structural conditioning signals for expert routing and factor construction.

\paragraph{Design rationale.}
The two stages are decoupled to avoid codebook collapse~\cite{van2017neural} and to enforce an information bottleneck. Stage~1 compresses cross-sectional variation into a stable discrete structure, while Stage~2 conditions temporal modeling on this representation without co-adapting the codebook.\footnote{Additional architectural details and shared neural building blocks are provided in the Technical Appendix~\ref{app:arch_impl}.}


\section{Experiments and Results}
\label{sec:experiments}

We evaluate PRISM-VQ on two major stock markets and conduct a comprehensive empirical study to assess its effectiveness, robustness, and interpretability.
All reported results are computed on a strictly out-of-sample test period spanning 2022 to 2024, using identical data preprocessing pipelines and train-validation-test splits across all methods.

\paragraph{Research Questions.}
Our experiments are designed to address the following questions:
\begin{itemize}
    \item \textbf{RQ1 (Overall effectiveness).} Does PRISM-VQ improve cross-sectional rank prediction and portfolio performances compared to strong baseline methods?
    \item \textbf{RQ2 (Component validity).} How do the three core components, namely expert prior factors, the discrete codebook, and the structure-conditioned MoE, individually contribute to overall performance?
    \item \textbf{RQ3 (Robustness).} How sensitive is PRISM-VQ to key hyperparameters, including codebook size, number of experts, and portfolio top-$K$ selection?
    \item \textbf{RQ4 (Interpretability).} Do the learned discrete codes capture economically meaningful structure beyond standard factor taxonomies?
    \item \textbf{RQ5 (Expert specialization).} Do different experts specialize in distinct market conditions while maintaining stable behavior over time?
\end{itemize}

\subsection{Experimental Setting}
\label{sec:exp_setting}

\subsubsection{Datasets and Protocol}
We evaluate PRISM-VQ on two major stock markets, CSI~300 (China) and S\&P~500 (U.S.), using data sourced from Qlib~\cite{yang2020qlib}. At each trading date $t$, we use contemporaneous index constituents to avoid survivorship bias. Input features are \texttt{Alpha158} factors with a lookback window of $T=20$ days.
The prediction target is the 5-day forward return, defined as $y_{i,t} = (\text{price}_{i,t+5}-\text{price}_{i,t+1})/\text{price}_{i,t+1}$, using the open price at $t+1$ and the close price at $t+5$. We additionally include 1- to 9-day forward returns as auxiliary targets for $\mathcal{L}_{\text{pred}}$. Expert prior factors consist of $P{=}13$ factor returns from the JKP Global Factor Library~\cite{jensen2023there}, computed using information available up to $t-1$ and aggregated as 20-day cumulative values.\footnote{Details on factor construction and preprocessing are provided in Technical Appendix~\ref{app:prior_factors}.}

We adopt chronological splits with training from 2009 to 2019, validation from 2020 to 2021, and testing from 2022 to 2024. Features are normalized using \texttt{RobustZScoreNorm}, missing values are forward-filled with \texttt{Fillna}, and training targets are cross-sectionally rank-normalized per date using \texttt{CSRankNorm}.

\subsubsection{Baselines}
We compare PRISM-VQ against representative baselines from three categories.
\textbf{General machine learning methods} include XGBoost~\cite{chen2016xgboost}, GRU~\cite{chung2014empirical}, TCN~\cite{lea2017temporal}, and Transformer~\cite{vaswani2017attention}.
\textbf{Factor-based methods} include CAE~\cite{gu2021autoencoder}, VAE~\cite{duan2022factorvae}, and VQVAE~\cite{kim2025factorvqvae}.
\textbf{Architecture-centric deep learning methods} include DTML~\cite{yoo2021accurate}, MASTER~\cite{li2024master}, and MATCC~\cite{cao2024matcc}. All baselines use identical data splits and preprocessing pipelines, with hyperparameters tuned on the validation set.

\subsubsection{Evaluation Metrics}
We evaluate performance using two complementary criteria. First, cross-sectional ranking accuracy is measured by daily RankIC and RankICIR. Second, portfolio performance is evaluated using a Top$K$-Drop$N$ strategy with $K{=}30$ and $N{=}5$, reporting annualized return (AR), maximum drawdown (MDD), and Sharpe ratio.\footnote{Formal definitions of ranking and portfolio evaluation metrics are given in Technical Appendix~\ref{app:metrics}, while details of the portfolio construction protocol follow Technical Appendix~\ref{app:topk}.} All portfolio results incorporate transaction costs of 5 basis points at open and 15 basis points at close.


\subsubsection{Implementation Details}
PRISM-VQ is implemented in PyTorch~\cite{paszke2019pytorch} and trained on a single NVIDIA RTX~3090 GPU. All experiments are repeated with five random seeds. In Stage~1, we use a representation dimension $d_s{=}128$ with a GRU encoder, a VQ codebook of size $K{=}512$, and a cross-asset Transformer with one layer and two attention heads. This stage is trained for up to 50 epochs. In Stage~2, we use a temporal dimension $d_t{=}64$ and a temporal Transformer with RoPE, using two heads for CSI~300 and four heads for S\&P~500, with dropout set to 0.1. The Transformer is followed by an MoE of lightweight MLP experts, with $M_e{=}2$ (top-1) for CSI~300 and $M_e{=}8$ (top-4) for S\&P~500. Stage~2 is trained for up to 50 epochs with the codebook fixed. Optimization is performed using AdamW~\cite{loshchilov2017decoupled} with a learning rate $10^{-4}$, LambdaLR scheduling, gradient clipping at 1.0, and early stopping with patience 15. We set $\lambda_{\text{commit}}{=}0.25$, $\lambda_{\text{contra}}{=}1$, $\lambda_{\text{pred}}{=}10^{-4}$, and $\lambda_{\text{balance}}{=}10^{-2}$ for CSI~300 and $10^{-3}$ for S\&P~500.


\subsection{Main Results (RQ1)}
\label{sec:main_results}

\begin{table}[t]
  \centering
  \caption{\textbf{Overall performance on CSI~300 and S\&P~500.}}
    \label{tab:overall}
  \footnotesize
  \setlength{\tabcolsep}{4pt}
  \renewcommand{\arraystretch}{1.12}
  \begin{adjustbox}{max width=\linewidth, center}
  \begin{tabular}{l l c c c c c}
    \toprule
    Market & Model & RankIC$\uparrow$ & RankICIR$\uparrow$ & AR$\uparrow$ & MDD$\downarrow$ & SR$\uparrow$ \\
    \midrule
    \multirow{11}{*}{\textbf{CSI~300}}
      & XGB$^{***}$      & 0.0496 & 0.3363 & 0.2430 & 0.2353 & 1.1549 \\
      & TCN$^{**}$       & 0.0518 & 0.3160 & 0.2281 & 0.2118 & 1.1879 \\
      & GRU              & 0.0590 & 0.3756 & 0.2347 & 0.2288 & \underline{1.2221} \\ 
      & Trans$^{***}$    & 0.0495 & 0.3100 & 0.2384 & 0.2478 & 1.1339 \\
      & CAE$^{***}$      & 0.0444 & 0.2843 & 0.2194 & 0.2322 & 1.0393   \\
      & VAE$^{**}$       & 0.0502 & 0.3360 & 0.1811 & \textbf{0.1906} & 1.0265 \\
      & VQVAE$^{*}$      & 0.0552 & 0.3641 & 0.2129 & 0.2074 & 1.0645 \\
      & DTML             & \underline{0.0625} & \underline{0.4076} & 0.2589 & 0.2069 & 1.1228   \\
      & MASTER$^{***}$   & 0.0401 & 0.2453 & 0.2193 & 0.2481 & 0.9101 \\
      & MATCC$^{***}$    & 0.0493 & 0.3295 & 0.2316 & 0.2299 & 1.0656 \\
      & PRISM-VQ         & \textbf{0.0646} & \textbf{0.4224} & \textbf{0.3077} & \underline{0.1924} & \textbf{1.5694} \\
    \midrule
    \multirow{11}{*}{\textbf{S\&P~500}}
      & XGB$^{***}$      & 0.0077 & 0.0713 & 0.1088 & 0.2079 & 0.4323 \\
      & TCN$^{***}$      & 0.0084 & \underline{0.0722} & 0.0880 & 0.2528 & 0.3841 \\
      & GRU$^{***}$      & 0.0057 & 0.0445 & 0.0867 & 0.3123 & 0.3425 \\
      & Trans$^{***}$    & 0.0054 & 0.0436 & 0.1004 & 0.2634 & 0.4025 \\
      & CAE$^{***}$      & 0.0063 & 0.0527 & 0.0825 & 0.2104 & 0.3952   \\
      & VAE$^{***}$      & 0.0043 & 0.0402 & 0.0734 & 0.2571 & 0.3702 \\
      & VQVAE$^{***}$    & 0.0046 & 0.0346 & 0.1109 & 0.1868 & \underline{0.5370} \\
      & DTML             & \underline{0.0089} & 0.0570 & 0.0667  & 0.3145  & 0.2726   \\
      & MASTER$^{***}$   & 0.0051 & 0.0321 & 0.1084 & \underline{0.1861} & 0.4662 \\
      & MATCC$^{***}$    & 0.0061 & 0.0478 & \underline{0.1147} & 0.2283 & 0.5164 \\
      & PRISM-VQ         & \textbf{0.0141} & \textbf{0.1208} & \textbf{0.1442} & \textbf{0.1616} & \textbf{0.6701} \\
    \bottomrule
    \end{tabular}
  \end{adjustbox}
  \vspace{0.5ex}
  \parbox{\linewidth}{
    \scriptsize \textit{Note}: RankIC and RankICIR are averaged over 5 random seeds; AR, MDD, and SR are computed from portfolios constructed using ensemble predictions. Stars indicate whether PRISM-VQ significantly outperforms each baseline on daily RankIC (block bootstrap, 10K resamples): $^{***}p<0.001$, $^{**}p<0.01$, $^{*}p<0.05$.
  }
\end{table}
%

\begin{figure}[t]
    \centering
    \includegraphics[width=\linewidth]{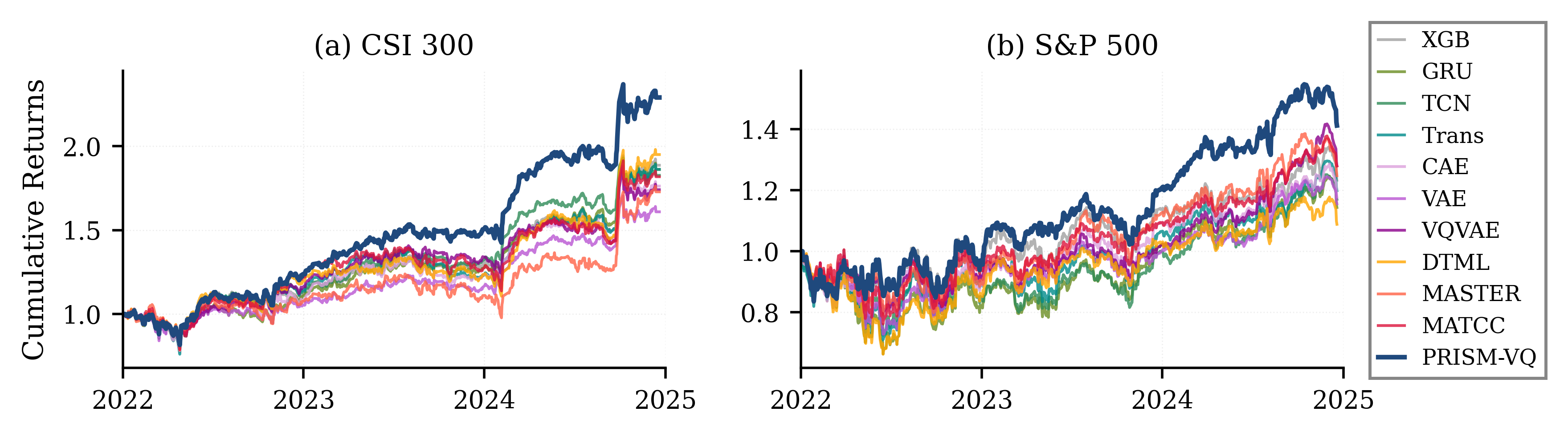}
    \caption{Cumulative returns (2022--2024). PRISM-VQ consistently outperforming baselines.}
    \label{fig:cumulative}
\end{figure}

Table~\ref{tab:overall} summarizes performance on CSI~300 and S\&P~500 over the 2022--2024 test period. On CSI~300, PRISM-VQ achieves a RankIC of 0.0646 and an SR of 1.57, outperforming all baselines, including DTML and GRU. It also attains the highest annualized return with a low MDD. On S\&P~500, PRISM-VQ reaches a RankIC of 0.0141 and an SR of 0.67, with the lowest MDD of 0.1616, representing a 58.4\% RankIC improvement over DTML. Statistical significance is evaluated on daily RankIC using a block bootstrap test with 10K resamples. PRISM-VQ significantly outperforms competing methods in 17 of 20 baseline comparisons across both markets at the $p<0.05$ level. These results demonstrate consistent improvements in ranking accuracy and portfolio performance across market regimes, as further illustrated by cumulative returns in Figure~\ref{fig:cumulative}.

\subsection{Ablation Study (RQ2)}

\begin{table}[t]
\centering
\caption{\textbf{Ablation study on key model components.}}
\label{tab:ablation}
\scriptsize
\setlength{\tabcolsep}{3pt}
\renewcommand{\arraystretch}{0.95}
\begin{tabular}{l cccc}
\toprule
& \multicolumn{2}{c}{CSI~300} & \multicolumn{2}{c}{S\&P~500} \\
\cmidrule(lr){2-3} \cmidrule(lr){4-5}
Configuration & RankIC & RankICIR & RankIC & RankICIR \\
\midrule
Full Model        & \textbf{0.0646} & \textbf{0.4224} & \textbf{0.0141} & \textbf{0.1208} \\
w/o Prior         & 0.0548 & 0.3756 & 0.0031 & 0.0314 \\
w/o MoE           & 0.0586 & 0.3958 & 0.0081 & 0.0736 \\
w/o Codebook      & 0.0472 & 0.2938 & $-$0.0024 & $-$0.0179 \\

\bottomrule
\end{tabular}
\end{table}

Table~\ref{tab:ablation} evaluates the contribution of the three core components of PRISM-VQ: expert prior factors ($f_p$), structure-conditioned MoE routing, and the discrete codebook. Removing the codebook causes the largest degradation, with RankIC dropping by 26.9\% on CSI~300 and becoming negative on S\&P~500, highlighting the importance of discrete latent structure. Excluding prior factors reduces RankIC by 15.2\% and 78.0\% on CSI~300 and S\&P~500, respectively, showing the regularizing role of expert knowledge. Replacing the MoE with a single network lowers RankIC by 9.3\% and 42.6\%, confirming the necessity of structure-conditioned temporal modeling. The consistently larger degradation on S\&P~500 indicates that all three components act synergistically to extract weak signals in efficient markets, whereas CSI~300’s stronger signal environment allows for partial compensation.

\subsection{Robustness Analysis(RQ3)}

\begin{figure}[t]
    \centering
    \includegraphics[width=0.9\linewidth]{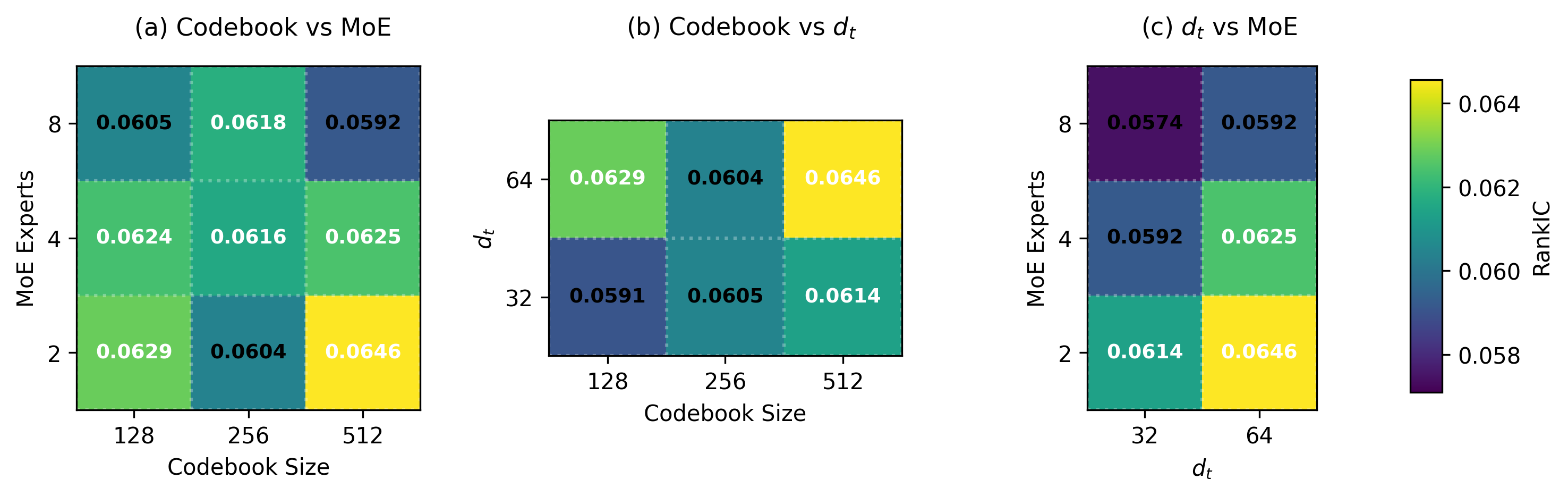}
    \includegraphics[width=0.9\linewidth]{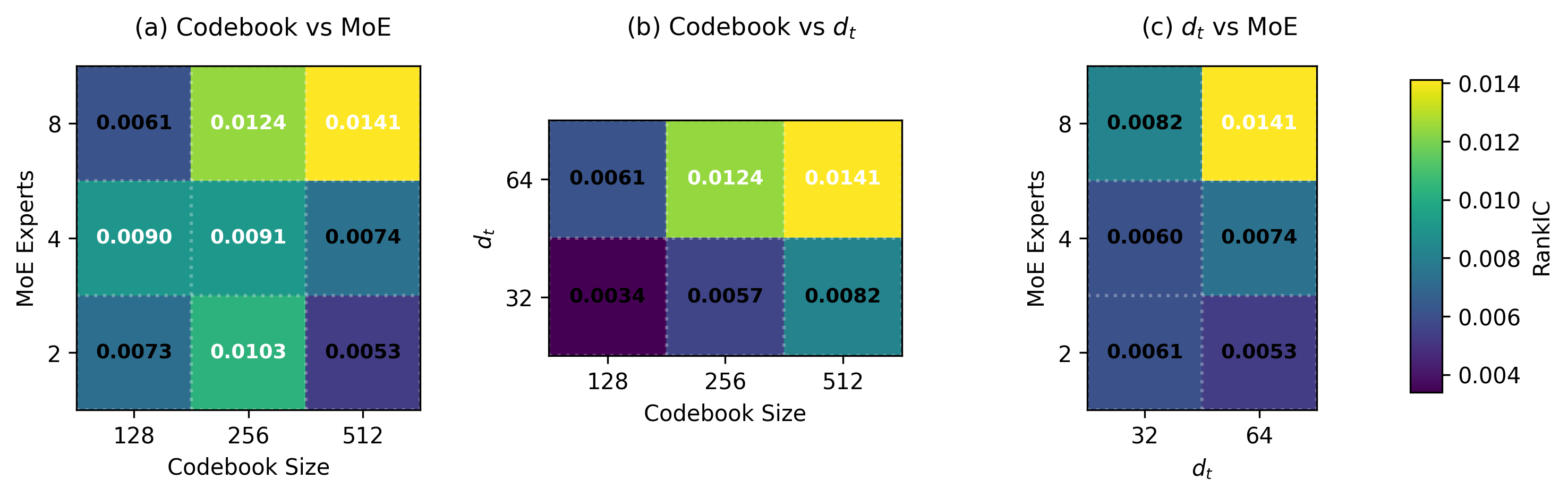}
    \caption{\textbf{Hyperparameter sensitivity analysis.}
    RankIC surfaces on CSI~300 (top) and S\&P~500 (bottom) as functions of codebook size $K$, temporal dimension $d_t$, and number of experts $M_e$.}
    \label{fig:sensitivity}
\end{figure}

\paragraph{Model hyperparameters.}
Figure~\ref{fig:sensitivity} shows sensitivity to the codebook size $K$, temporal dimension $d_t$, and number of experts $M_e$. Across both markets, performance peaks at $d_t=64$ and $K=512$, indicating that moderate temporal capacity with fine-grained quantization is optimal. The optimal number of experts differs by market: CSI~300 peaks at $M_e=2$, while S\&P~500 improves up to $M_e=8$, suggesting that more efficient markets benefit from greater expert specialization. The shared optimum at $K=512$ indicates that fine-grained discrete quantization is broadly beneficial.

\paragraph{Portfolio size.}
\label{robust:port_size}
Table~\ref{tab:portfolio_k} reports performance for larger Top-$K$ portfolios ($K\in\{40,50\}$). PRISM-VQ consistently outperforms all baselines across portfolio sizes, achieving an AR of 27.85\% and 25.35\% on CSI~300 and 13.96\% and 12.42\% on S\&P~500 for $K{=}40$ and $K{=}50$, respectively, representing improvements of 17--27\% over the strongest baselines. PRISM-VQ also attains the lowest MDD on S\&P~500 across all settings (15.82--16.07\%), while gradual performance degradation with larger $K$ indicates robust signals beyond top-ranked stocks.\footnote{Additional robustness under varying transaction costs is reported in Technical Appendix~\ref{app:tcrobust}.}

\begin{table}[t]
\centering
\caption{\textbf{Portfolio performance across Top-$K$ sizes.}} 
\label{tab:portfolio_k}
\tiny
\setlength{\tabcolsep}{2pt}
\renewcommand{\arraystretch}{0.9}
\begin{tabular}{l ccc ccc ccc ccc}
\toprule
& \multicolumn{6}{c}{CSI~300} & \multicolumn{6}{c}{S\&P~500} \\
\cmidrule(lr){2-7} \cmidrule(lr){8-13}
& \multicolumn{3}{c}{$K$=40} & \multicolumn{3}{c}{$K$=50} & \multicolumn{3}{c}{$K$=40} & \multicolumn{3}{c}{$K$=50} \\
\cmidrule(lr){2-4} \cmidrule(lr){5-7} \cmidrule(lr){8-10} \cmidrule(lr){11-13}
Model & AR & MDD & SR & AR & MDD & SR & AR & MDD & SR & AR & MDD & SR \\
\midrule
XGB      & 0.22 & 0.22 & 1.11 & 0.19 & 0.22 & 0.97 & 0.11 & 0.20 & 0.45 & 0.10 & 0.20 & 0.42 \\
GRU      & 0.20 & 0.23 & 1.09 & 0.19 & 0.23 & 1.05 & 0.08 & 0.30 & 0.35 & 0.07 & 0.28 & 0.30 \\
TCN      & 0.22 & 0.21 & \underline{1.19} & \underline{0.21} & 0.21 & \underline{1.15} & 0.10 & 0.24 & 0.43 & 0.10 & 0.25 & 0.46 \\
Trans    & \underline{0.24} & 0.23 & 1.19 & 0.20 & 0.23 & 1.03 & 0.08 & 0.28 & 0.35 & 0.08 & 0.27 & 0.35 \\
CAE      & 0.03 & 0.21 & 0.17 & 0.04 & 0.22 & 0.19 & 0.08 & 0.21 & 0.42 & 0.08 & 0.20 & 0.44 \\
VAE      & 0.19 & \underline{0.19} & 1.08 & 0.17 & \underline{0.20} & 0.97 & 0.09 & 0.24 & 0.47 & \underline{0.11} & 0.22 & \underline{0.54} \\
VQVAE    & 0.20 & 0.19 & 1.02 & 0.17 & \textbf{0.19} & 0.91 & 0.09 & 0.20 & 0.45 & 0.09 & 0.21 & 0.47 \\
DTML     & 0.02 & 0.22 & 0.12 & 0.03 & 0.22 & 0.15 & 0.07 & 0.19 & 0.39 & 0.06 & 0.19 & 0.36 \\
MASTER   & 0.19 & 0.26 & 0.81 & 0.18 & 0.25 & 0.77 & 0.10 & 0.21 & 0.44 & 0.10 & 0.22 & 0.43 \\
MATCC    & 0.19 & 0.24 & 0.92 & 0.18 & 0.24 & 0.86 & \underline{0.11} & 0.21 & \underline{0.51} & 0.10 & 0.22 & 0.47 \\
PRISM-VQ & \textbf{0.28} & \textbf{0.18} & \textbf{1.48} & \textbf{0.25} & 0.20 & \textbf{1.37} & \textbf{0.14} & \textbf{0.16} & \textbf{0.67} & \textbf{0.12} & \textbf{0.16} & \textbf{0.61} \\
\bottomrule
\end{tabular}
\end{table}

\subsection{Interpretation of Representations (RQ4)}

\begin{figure}[t]
    \centering
    \includegraphics[width=\linewidth]{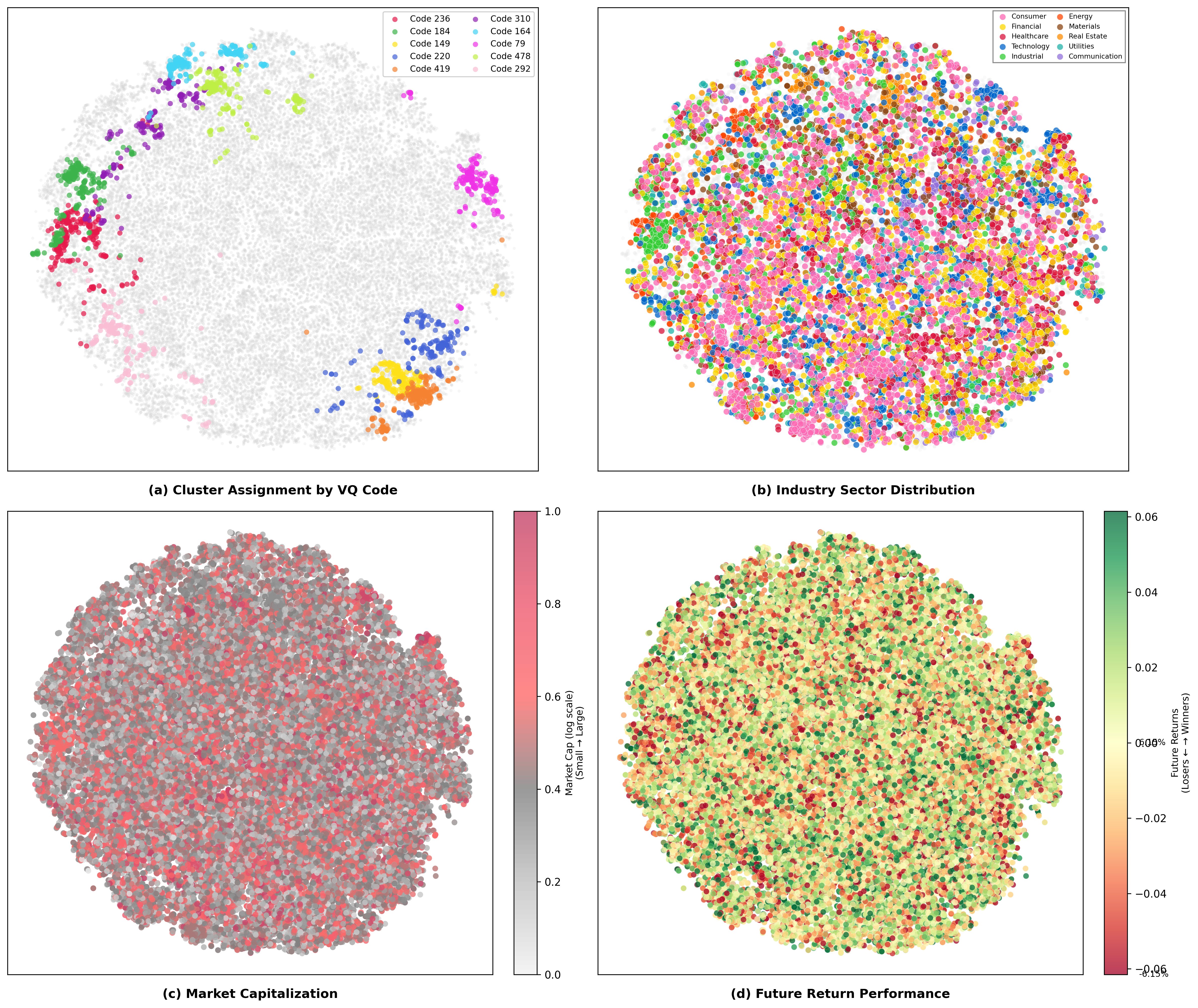}
    \caption{\textbf{t-SNE visualization of learned representations (S\&P~500, 2024).} (a) Top 10 most frequent codes. (b) 41 representative stocks colored by sector. (c) Color gradient by market capitalization. (d) Color gradient by forward returns.}
    \label{fig:tsne}
\end{figure}

\paragraph{Visualization of learned representations.}
Figure~\ref{fig:tsne} presents a t-SNE visualization of learned discrete representations on the S\&P~500 in 2024 with representative stocks.\footnote{The representative stocks are listed in Technical Appendix~\ref{app:stocks}.} In panel (a), the ten most frequent codes form well-separated clusters. In panel (b), stocks from different sectors are substantially mixed across clusters, indicating that vector quantization captures structure beyond sector taxonomies encoded by prior factors. Although panels (c) and (d) show that larger-cap and positive-return stocks tend to concentrate in certain regions, the lack of clear separation indicates that the learned codes are not driven solely by market capitalization or future returns. Instead, they capture a richer cross-sectional structure beyond size or return effects.

\paragraph{Factor exposure analysis.}
Economic interpretability is evaluated by computing Spearman correlations between code-specific returns and 13 JKP factor returns (Table~\ref{tab:factor_exposure}).
The learned codes exhibit systematic but moderate factor associations. For example, S\&P~500 Code~72 loads on profitability, low-risk, and momentum, consistent with a quality-growth profile. Maximum correlations remain modest (approximately 0.12), indicating that the codes capture multi-dimensional factor combinations rather than replicating individual factors. 
The codes also exhibit temporal stability, with monthly persistence rates of 4.9\% on CSI~300 and 7.4\% on S\&P~500, substantially exceeding the 1\% uniform baseline.\footnote{Further analysis is provided in Technical Appendix~\ref{app:transition}.}

\begin{table}[t]
    \caption{\textbf{Factor exposures of selected VQ codes.}}
    \label{tab:factor_exposure}
    \centering
    \scriptsize
    \setlength{\tabcolsep}{3pt}
    \begin{tabular}{cl rrr rrr}
        \toprule
        & & \multicolumn{3}{c}{Top 3 Factors} & \multicolumn{3}{c}{$\rho$} \\
        \cmidrule(lr){3-5} \cmidrule(lr){6-8}
        Market & Code & F1 & F2 & F3 & $\rho_1$ & $\rho_2$ & $\rho_3$ \\
        \midrule
        \multirow{3}{*}{S\&P}
        & 72  & Prof. & LowRisk & Mom. & +.118 & +.109 & +.096 \\
        & 34  & Qual. & Size & Prof. & $-$.113 & +.097 & $-$.077 \\
        & 224 & Prof. & LowLev. & LowRisk & $-$.113 & +.112 & $-$.110 \\
        \midrule
        \multirow{3}{*}{CSI}
        & 482 & Qual. & Accr. & Invest. & $-$.118 & +.115 & +.113 \\
        & 247 & LowLev. & Value & Invest. & $-$.117 & +.117 & +.103 \\
        & 179 & Size & Prof. & Qual. & +.117 & $-$.081 & $-$.077 \\
        \bottomrule
    \end{tabular}
\end{table}

\subsection{Expert Specialization (RQ5)}

\begin{figure*}[!htbp]
    \centering
    \includegraphics[width=0.45\textwidth]{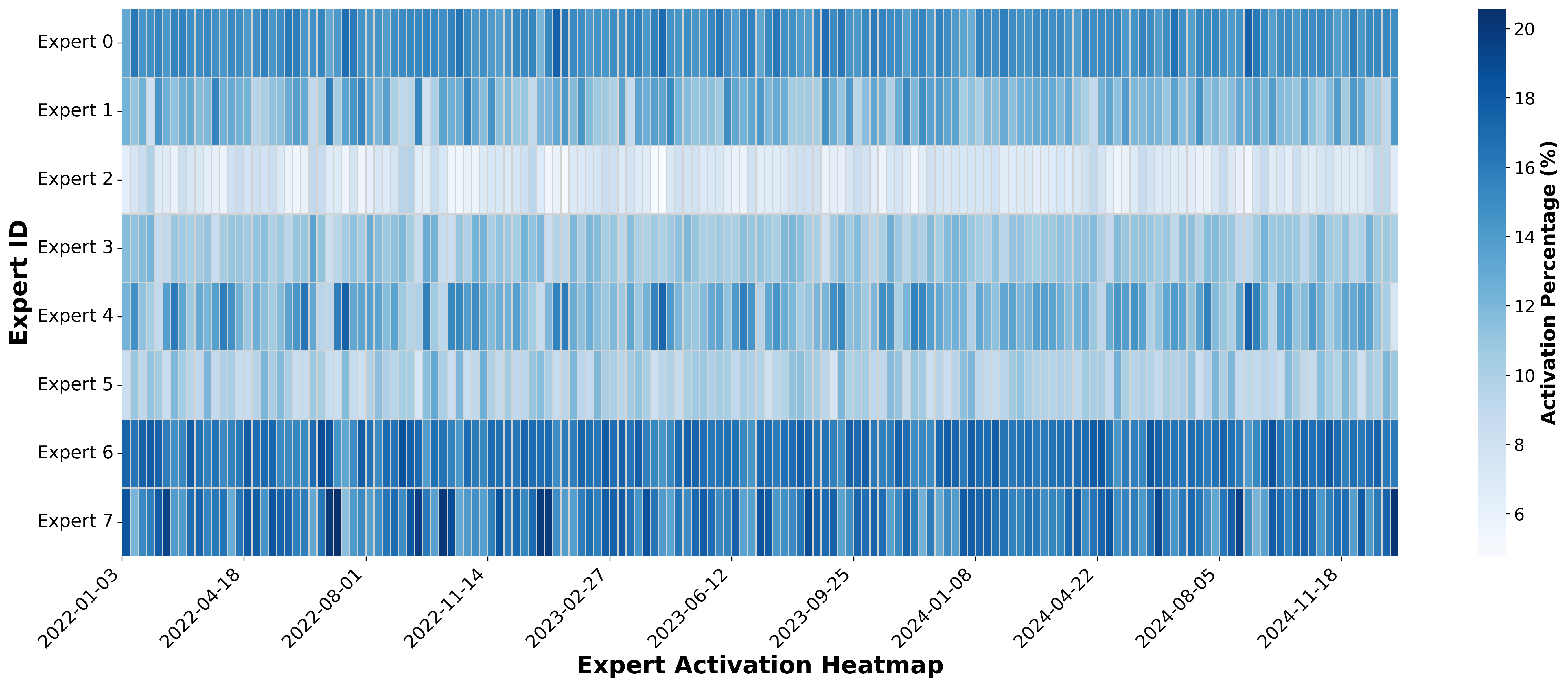}
    \includegraphics[width=0.49\textwidth]{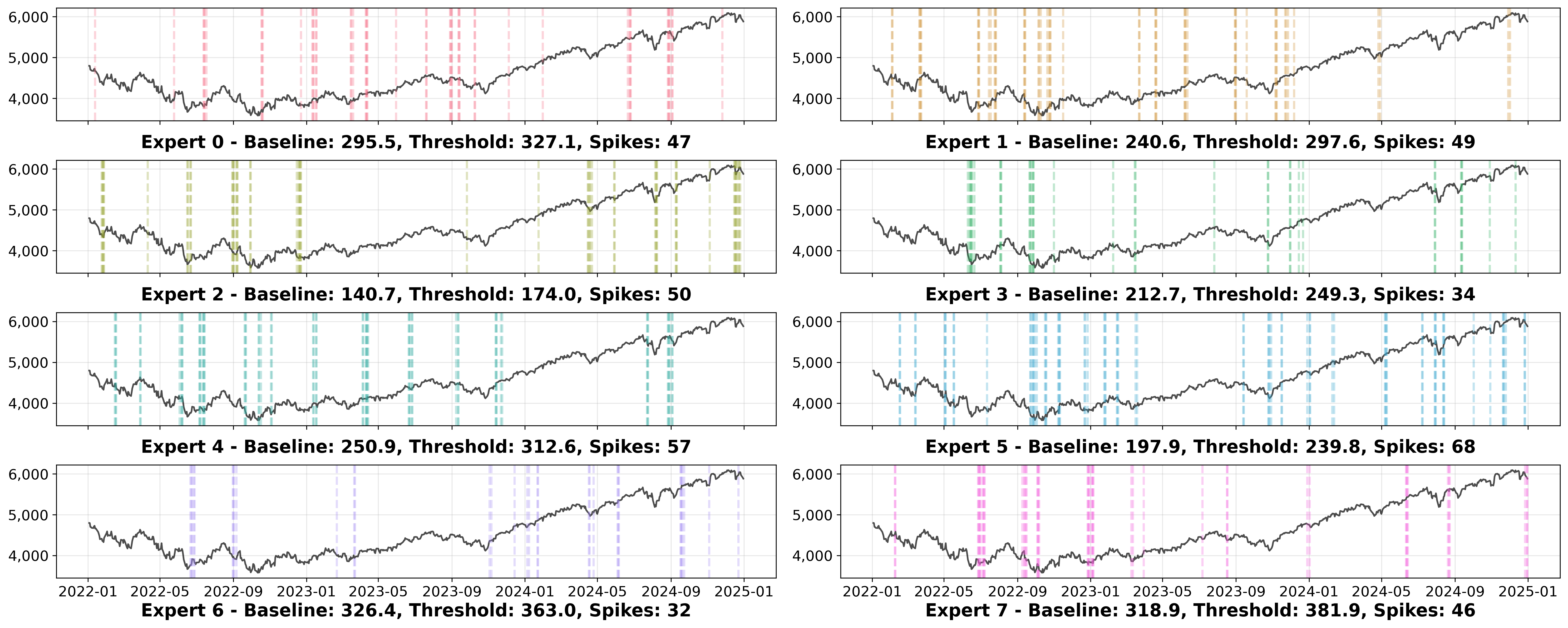}
    \caption{\textbf{MoE activation patterns (S\&P~500)}. Left: weekly mean activation, Right: activation spikes overlaid on market trajectory.}
    \label{fig:expert_activation}
\end{figure*} 

Figure~\ref{fig:expert_activation} shows the MoE routing on the S\&P~500. The left heatmap reveals heterogeneous expert utilization: Experts~6 and~7 dominate with approximately 16--20\% average activation, while Expert~2 remains sparse at around 5--7\%.

Regime-dependent timing is examined in the right panel, which visualizes daily activation spikes computed from the mean gating weight $P_{e,t}=\frac{1}{N_t}\sum_{i=1}^{N_t} G_{i,e,t}$, consistent with Eq.~\eqref{eq:temporal_obj}.
Spikes are identified when $P_{e,t}$ exceeds $\mu_e+1.5\sigma_e$. Pairwise Wilcoxon signed-rank tests across all 28 expert pairs indicate statistically distinct activation distributions, with all $p$-values below 0.001. The strongest separation occurs between Expert~2 (mean activation 7.09\%) and Expert~6 (mean activation 16.46\%). Expert~2 functions as a rare specialist that activates in specific market regimes, whereas Expert~6 remains consistently engaged.

Low overlap in spike dates further supports temporal specialization.
The Jaccard overlap is 6.8\%, compared to an 8.11\% baseline, indicating that experts respond to distinct market conditions rather than activating synchronously.


\section{Conclusion}
\label{sec:conclusion}

We proposed PRISM-VQ, a two-stage framework for cross-sectional stock rank prediction that learns reusable discrete cross-sectional structure via vector quantization with contrastive regularization and generates time-varying factor loadings through structure-conditioned MoE routing guided by financial priors. Across CSI~300 and S\&P~500, PRISM-VQ consistently improves prediction and portfolio performance over strong baselines while producing interpretable discrete codes and specialized expert behaviors. These results indicate that combining a discrete information bottleneck with prior-informed conditional computation is an effective design principle for low-SNR, non-stationary financial return prediction problems.




\section*{Acknowledgments}
This work was supported by the National Research Foundation of Korea (NRF) grant funded by the Korea government (MSIT) (No. RS-2025-24803415). The authors gratefully acknowledge the support of RiskX Corp.




\bibliographystyle{named}
\bibliography{refs}

@inproceedings{perez2018film,
  title={Film: Visual reasoning with a general conditioning layer},
  author={Perez, Ethan and Strub, Florian and De Vries, Harm and Dumoulin, Vincent and Courville, Aaron},
  booktitle={Proceedings of the AAAI conference on artificial intelligence},
  volume={32},
  number={1},
  year={2018}
}

@inproceedings{duan2025factorgcl,
  title={FactorGCL: A Hypergraph-Based Factor Model with Temporal Residual Contrastive Learning for Stock Returns Prediction},
  author={Duan, Yitong and Wang, Weiran and Li, Jian},
  booktitle={Proceedings of the AAAI Conference on Artificial Intelligence},
  volume={39},
  number={1},
  pages={173--181},
  year={2025}
}

@inproceedings{xiang2024rsap,
  title={Rsap-dfm: Regime-shifting adaptive posterior dynamic factor model for stock returns prediction},
  author={Xiang, Quanzhou and Chen, Zhan and Sun, Qi and Jiang, Rujun},
  booktitle={Proceedings of the Thirty-Third International Joint Conference on Artificial Intelligence, IJCAI-24. International Joint Conferences on Artificial Intelligence Organization},
  year={2024}
}

@inproceedings{yoo2021accurate,
  title={Accurate multivariate stock movement prediction via data-axis transformer with multi-level contexts},
  author={Yoo, Jaemin and Soun, Yejun and Park, Yong-chan and Kang, U},
  booktitle={Proceedings of the 27th ACM SIGKDD Conference on Knowledge Discovery \& Data Mining},
  pages={2037--2045},
  year={2021}
}

@inproceedings{du2024explainable,
  title={Explainable stock price movement prediction using contrastive learning},
  author={Du, Kelvin and Mao, Rui and Xing, Frank and Cambria, Erik},
  booktitle={Proceedings of the 33rd ACM International Conference on Information and Knowledge Management},
  pages={529--537},
  year={2024}
}

@article{zhao2024storm,
  title={STORM: A Spatio-Temporal Factor Model Based on Dual Vector Quantized Variational Autoencoders for Financial Trading},
  author={Zhao, Yilei and Zhang, Wentao and Yang, Tingran and Jiang, Yong and Huang, Fei and Lim, Wei Yang Bryan},
  journal={arXiv preprint arXiv:2412.09468},
  year={2024}
}

@inproceedings{hwang2023simstock,
  title={SimStock: Representation Model for Stock Similarities},
  author={Hwang, Yoontae and Lee, Junhyeong and Kim, Daham and Noh, Seunghwan and Hong, Joohwan and Lee, Yongjae},
  booktitle={Proceedings of the Fourth ACM International Conference on AI in Finance},
  pages={533--540},
  year={2023}
}

@inproceedings{cao2024matcc,
  title={Matcc: A novel approach for robust stock price prediction incorporating market trends and cross-time correlations},
  author={Cao, Zhiyuan and Xu, Jiayu and Dong, Chengqi and Yu, Peiwen and Bai, Tian},
  booktitle={Proceedings of the 33rd ACM International Conference on Information and Knowledge Management},
  pages={187--196},
  year={2024}
}

@inproceedings{duan2022factorvae,
  title={Factorvae: A probabilistic dynamic factor model based on variational autoencoder for predicting cross-sectional stock returns},
  author={Duan, Yitong and Wang, Lei and Zhang, Qizhong and Li, Jian},
  booktitle={Proceedings of the AAAI conference on artificial intelligence},
  volume={36},
  number={4},
  pages={4468--4476},
  year={2022}
}

@article{kim2025factorvqvae,
  title={FactorVQVAE: Discrete latent factor model via Vector Quantized Variational Autoencoder},
  author={Kim, Namhyoung and Ock, Seung Eun and Song, Jae Wook},
  journal={Knowledge-Based Systems},
  volume={318},
  pages={113460},
  year={2025},
  publisher={Elsevier}
}

@article{shazeer2017outrageously,
  title={Outrageously large neural networks: The sparsely-gated mixture-of-experts layer},
  author={Shazeer, Noam and Mirhoseini, Azalia and Maziarz, Krzysztof and Davis, Andy and Le, Quoc and Hinton, Geoffrey and Dean, Jeff},
  journal={arXiv preprint arXiv:1701.06538},
  year={2017}
}

@article{van2017neural,
  title={Neural discrete representation learning},
  author={Van Den Oord, Aaron and Vinyals, Oriol and others},
  journal={Advances in neural information processing systems},
  volume={30},
  year={2017}
}

@article{vaswani2017attention,
  title={Attention is all you need},
  author={Vaswani, Ashish and Shazeer, Noam and Parmar, Niki and Uszkoreit, Jakob and Jones, Llion and Gomez, Aidan N and Kaiser, {\L}ukasz and Polosukhin, Illia},
  journal={Advances in neural information processing systems},
  volume={30},
  year={2017}
}

@article{gu2021autoencoder,
  title={Autoencoder asset pricing models},
  author={Gu, Shihao and Kelly, Bryan and Xiu, Dacheng},
  journal={Journal of Econometrics},
  volume={222},
  number={1},
  pages={429--450},
  year={2021},
  publisher={Elsevier}
}

@article{sharpe1964capital,
  title={Capital asset prices: A theory of market equilibrium under conditions of risk},
  author={Sharpe, William F},
  journal={The journal of finance},
  volume={19},
  number={3},
  pages={425--442},
  year={1964},
  publisher={Wiley Online Library}
}

@inproceedings{chen2016xgboost,
  title={Xgboost: A scalable tree boosting system},
  author={Chen, Tianqi and Guestrin, Carlos},
  booktitle={Proceedings of the 22nd acm sigkdd international conference on knowledge discovery and data mining},
  pages={785--794},
  year={2016}
}

@article{fama1992cross,
  title={The cross-section of expected stock returns},
  author={Fama, Eugene F and French, Kenneth R},
  journal={the Journal of Finance},
  volume={47},
  number={2},
  pages={427--465},
  year={1992},
  publisher={Wiley Online Library}
}

@article{chung2014empirical,
  title={Empirical evaluation of gated recurrent neural networks on sequence modeling},
  author={Chung, Junyoung and Gulcehre, Caglar and Cho, KyungHyun and Bengio, Yoshua},
  journal={arXiv preprint arXiv:1412.3555},
  year={2014}
}

@article{cochrane2011presidential,
  title={Presidential address: Discount rates},
  author={Cochrane, John H},
  journal={The Journal of finance},
  volume={66},
  number={4},
  pages={1047--1108},
  year={2011},
  publisher={Wiley Online Library}
}

@article{fama1993common,
  title={Common risk factors in the returns on stocks and bonds},
  author={Fama, Eugene F and French, Kenneth R},
  journal={Journal of financial economics},
  volume={33},
  number={1},
  pages={3--56},
  year={1993},
  publisher={Elsevier}
}

@inproceedings{lea2017temporal,
  title={Temporal convolutional networks for action segmentation and detection},
  author={Lea, Colin and Flynn, Michael D and Vidal, Rene and Reiter, Austin and Hager, Gregory D},
  booktitle={proceedings of the IEEE Conference on Computer Vision and Pattern Recognition},
  pages={156--165},
  year={2017}
}

@article{yang2020qlib,
  title={Qlib: An ai-oriented quantitative investment platform},
  author={Yang, Xiao and Liu, Weiqing and Zhou, Dong and Bian, Jiang and Liu, Tie-Yan},
  journal={arXiv preprint arXiv:2009.11189},
  year={2020}
}

@inproceedings{li2024master,
  title={MASTER: Market-Guided Stock Transformer for Stock Price Forecasting},
  author={Li, Tong and Liu, Zhaoyang and Shen, Yanyan and Wang, Xue and Chen, Haokun and Huang, Sen},
  booktitle={Proceedings of the AAAI Conference on Artificial Intelligence},
  volume={38},
  number={1},
  pages={162--170},
  year={2024}
}

@inproceedings{chen2020simple,
  title={A simple framework for contrastive learning of visual representations},
  author={Chen, Ting and Kornblith, Simon and Norouzi, Mohammad and Hinton, Geoffrey},
  booktitle={International conference on machine learning},
  pages={1597--1607},
  year={2020},
  organization={PmLR}
}

@article{paszke2019pytorch,
  title={Pytorch: An imperative style, high-performance deep learning library},
  author={Paszke, Adam and Gross, Sam and Massa, Francisco and Lerer, Adam and Bradbury, James and Chanan, Gregory and Killeen, Trevor and Lin, Zeming and Gimelshein, Natalia and Antiga, Luca and others},
  journal={Advances in neural information processing systems},
  volume={32},
  year={2019}
}

@article{loshchilov2017decoupled,
  title={Decoupled weight decay regularization},
  author={Loshchilov, Ilya and Hutter, Frank},
  journal={arXiv preprint arXiv:1711.05101},
  year={2017}
}

@article{su2024roformer,
  title={Roformer: Enhanced transformer with rotary position embedding},
  author={Su, Jianlin and Ahmed, Murtadha and Lu, Yu and Pan, Shengfeng and Bo, Wen and Liu, Yunfeng},
  journal={Neurocomputing},
  volume={568},
  pages={127063},
  year={2024},
  publisher={Elsevier}
}

@article{israel2020can,
  title={Can Machines' Learn'Finance?},
  author={Israel, Ronen and Kelly, Bryan T and Moskowitz, Tobias J},
  journal={Journal of Investment Management},
  year={2020}
}

@article{zhang2025major,
  title={Major issues in high-frequency financial data analysis: A survey of solutions},
  author={Zhang, Lu and Hua, Lei},
  journal={Mathematics},
  volume={13},
  number={3},
  pages={347},
  year={2025},
  publisher={MDPI}
}

@article{jensen2023there,
  title={Is there a replication crisis in finance?},
  author={Jensen, Theis Ingerslev and Kelly, Bryan and Pedersen, Lasse Heje},
  journal={The Journal of Finance},
  volume={78},
  number={5},
  pages={2465--2518},
  year={2023},
  publisher={Wiley Online Library}
}

@inproceedings{chen2024automatic,
  title={Automatic de-biased temporal-relational modeling for stock investment recommendation},
  author={Chen, Weijun and Li, Shun and Yu, Xipu and Wang, Heyuan and Chen, Wei and Wang, Tengjiao},
  booktitle={Proceedings of the Thirty-Third International Joint Conference on Artificial Intelligence},
  pages={1999--2008},
  year={2024}
}

@article{song2025multi,
  title={Multi-Scale Temporal Neural Network for Stock Trend Prediction Enhanced by Temporal Hyepredge Learning},
  author={Song, Lingyun and Li, Haodong and Chen, Siyu and Gan, Xinbiao and Shi, Binze and Ma, Jie and Pan, Yudai and Wang, Xiaoqi and Shang, Xuequn},
  journal={Proceedings of the IJCAI, Montreal, QC, Canada},
  pages={16--22},
  year={2025}
}

@inproceedings{alaygut2025hypergraph,
  title={Hypergraph neural networks to predict stock movements by exploring higher-order relationships},
  author={Alaygut, Tuna and Sefer, Emre},
  booktitle={Proceedings of the 6th ACM International Conference on AI in Finance},
  pages={700--708},
  year={2025}
}

@inproceedings{kim2021reversible,
  title={Reversible instance normalization for accurate time-series forecasting against distribution shift},
  author={Kim, Taesung and Kim, Jinhee and Tae, Yunwon and Park, Cheonbok and Choi, Jang-Ho and Choo, Jaegul},
  booktitle={International conference on learning representations},
  year={2021}
}

\clearpage

\appendix

\nolinenumbers
\counterwithin*{equation}{section}
\renewcommand\theequation{\thesection.\arabic{equation}}
\counterwithin*{figure}{section}
\renewcommand\thefigure{\thesection.\arabic{figure}}
\counterwithin*{table}{section}
\renewcommand\thetable{\thesection.\arabic{table}}
\counterwithin*{algorithm}{section}
\renewcommand\thealgorithm{\thesection.\arabic{algorithm}}

\twocolumn[
  \begin{center}
    \LARGE \bf Technical Appendix of Vector-Quantized Discrete Latent Factors Meet Financial Priors: Dynamic Cross-Sectional Stock Ranking Prediction for Portfolio Construction
  \end{center}
  \bigskip
  \begin{center}
    \large \bf Namhyoung Kim$^{1,2}$ \quad Jae Wook Song$^2$ \\ \vspace{1mm}
    \large \textnormal{$^1$RiskX \quad $^2$Hanyang University} \\ \vspace{1mm}
    \large \textnormal{namhyoung.kim@riskx.tech, \quad jwsong@hanyang.ac.kr}
  \end{center}
  \vspace{2\baselineskip} 
]
\normalsize

This appendix provides detailed architectural and implementation descriptions of PRISM-VQ. The goal is to complement the main text by making all modeling choices explicit and reproducible. 

\section{Architecture and Implementation Details}
\label{app:arch_impl}

We first describe the neural building blocks shared across model components and then explain how they are instantiated within each learning stage.

Throughout this appendix, we follow the notation of the main text. Stage~1 (spatial learning) employs a GRU encoder $\phi_{\text{gru}}$, a cross-asset Transformer encoder $\mathcal{E}$, and a vector-quantization (VQ) codebook $\mathcal{Z}$ to produce discrete stock representations $z_{q,i}\in\mathbb{R}^{d_s}$. Stage~2 (temporal learning) uses a temporal Transformer $\mathcal{T}$ together with a structure-conditioned Mixture-of-Experts (MoE) to generate dynamic factor loadings $\boldsymbol{\beta}_i=(\alpha_i,\beta_{p,i},\beta_{l,i})$.

Unless stated otherwise, experiments use a lookback window of $T{=}20$ trading days, $C{=}158$ Alpha158 input features, $P{=}13$ expert prior factors, a spatial latent dimension of $d_s{=}128$, a temporal hidden dimension of $d_t{=}64$, and a VQ codebook of size $K{=}512$.

\subsection{Shared Model Building Blocks}
\label{app:building_blocks}

This subsection describes the neural components reused across both stages of PRISM-VQ. These include Transformer encoder blocks for modeling cross-sectional and temporal dependencies, as well as the positional encoding mechanism used for sequential inputs.

\subsubsection{Transformer Encoder Block}
\label{app:transformer_block}

Let $X \in \mathbb{R}^{N \times d}$ denote a matrix of $N$ input tokens with a hidden dimension of $d$. In Stage~1, tokens correspond to stocks in the cross section at a fixed time. In Stage~2, tokens correspond to temporal observations for a single stock. Each Transformer encoder block consists of a multi-head self-attention (MHA) layer followed by a position-wise feed-forward network (FFN), with residual connections and layer normalization applied after each sub-layer.
This architecture enables flexible modeling of interactions while maintaining stable optimization.

We use $n_h$ attention heads, each with a per-head dimension of $d_h = d / n_h$.

\paragraph{Multi-Head Self-Attention (MHA).}
For each attention head $h \in \{1,\ldots,n_h\}$, queries, keys, and values are computed via linear projections:
\begin{align}
Q^{(h)} = X W_Q^{(h)},\quad
K^{(h)} = X W_K^{(h)},\quad
V^{(h)} = X W_V^{(h)}, \nonumber\\
W_Q^{(h)}, W_K^{(h)}, W_V^{(h)} \in \mathbb{R}^{d \times d_h},\quad
Q^{(h)}, K^{(h)}, V^{(h)} \in \mathbb{R}^{N \times d_h}.
\label{eq:mha_proj}
\end{align}

Scaled dot-product attention computes pairwise interactions between tokens:
\begin{align}
A^{(h)} &= \mathrm{softmax}\!\left(\frac{Q^{(h)} {K^{(h)}}^{\top}}{\sqrt{d_h}}\right) \in \mathbb{R}^{N \times N}, \nonumber\\
\mathrm{head}^{(h)} &= A^{(h)} V^{(h)} \in \mathbb{R}^{N \times d_h}.
\label{eq:sdpa}
\end{align} The scaling by $\sqrt{d_h}$ prevents attention scores from growing excessively large and stabilizes gradients.

Outputs from all heads are concatenated and projected back to dimension $d$:
\begin{align}
\mathrm{MHA}(X)
&= \mathrm{Concat}\big(\mathrm{head}^{(1)},\ldots,\mathrm{head}^{(n_h)}\big)\, W_O, \nonumber\\
&\qquad W_O \in \mathbb{R}^{d \times d}.
\label{eq:mha_out}
\end{align}

\paragraph{Feed-Forward Network (FFN).}
The FFN applies a position-wise nonlinear transformation independently to each token:
\begin{align}
\mathrm{FFN}(x)
&= W_2\,\mathrm{GELU}(W_1 x + b_1) + b_2, \nonumber\\
&\qquad W_1 \in \mathbb{R}^{d \times d_{\mathrm{ff}}},\;
W_2 \in \mathbb{R}^{d_{\mathrm{ff}} \times d},
\label{eq:ffn}
\end{align} where $d_{\mathrm{ff}}$ denotes the intermediate expansion dimension, typically chosen as a multiple of $d$.
The GELU nonlinearity provides smooth activation and empirically improves optimization stability in Transformers.

\paragraph{Encoder block structure.}
Each Transformer encoder block combines the above components with residual connections and layer normalization:
\begin{align}
\tilde{X} &= \mathrm{LayerNorm}\!\big(X + \mathrm{MHA}(X)\big), \nonumber\\
X' &= \mathrm{LayerNorm}\!\big(\tilde{X} + \mathrm{FFN}(\tilde{X})\big).
\label{eq:encoder_block}
\end{align}

Residual connections preserve information flow across layers, while layer normalization mitigates covariate shift and improves training stability.

\subsubsection{Rotary Position Embeddings (RoPE)}
\label{app:rope}

To encode temporal order in sequential inputs, PRISM-VQ adopts Rotary Position Embeddings (RoPE)~\cite{su2024roformer}.
Unlike absolute positional encodings, RoPE injects relative positional information directly into the attention mechanism, allowing attention scores to depend on the relative distances between time steps.

RoPE is applied independently to each attention head and operates on the per-head dimension $d_h$.
For a query or key vector $u \in \mathbb{R}^{d_h}$ at position index $m$, RoPE applies a sequence of two-dimensional rotations:
\begin{align}
\mathrm{RoPE}(u,m) = \mathrm{BlockDiag}(R_{m,1},\ldots,R_{m,d_h/2})\,u,
\label{eq:rope}
\end{align}
where each rotation matrix is defined as
\begin{align}
R_{m,j}
&=
\begin{pmatrix}
\cos(m\theta_j) & -\sin(m\theta_j)\\
\sin(m\theta_j) & \cos(m\theta_j)
\end{pmatrix}, \nonumber\\
\theta_j &= 10000^{-2(j-1)/d_h},
\qquad j\in\{1,\ldots,d_h/2\}.
\label{eq:rope_block}
\end{align}

In practice, the RoPE transformation replaces the original $Q^{(h)}$ and $K^{(h)}$ vectors in Eq.~\eqref{eq:sdpa} before attention computation.
This enables the model to generalize to unseen sequence lengths and improves robustness when modeling non-stationary financial time series.

\subsection{Two-Stage Training: What is Learned vs.\ Fixed}
\label{app:two_stage_freeze}

PRISM-VQ is trained using a two-stage procedure designed to decouple cross-sectional structure discovery from temporal modeling. Table~\ref{tab:freeze} summarizes which parameters are optimized in each stage. After Stage~1, the learned VQ codebook $\mathcal{Z}$ is frozen. At each date, discrete assignments are recomputed by nearest-neighbor search in the fixed codebook, allowing stock-level codes to vary over time while preserving a stable set of structural prototypes.

\begin{table}[t]
\centering
\caption{Learned vs.\ fixed parameters in two-stage training.}
\label{tab:freeze}
\small
\setlength{\tabcolsep}{6pt}
\renewcommand{\arraystretch}{1.1}
\begin{adjustbox}{width=\columnwidth}
    \begin{tabular}{lcc}
    \toprule
    Module & Stage~1 training & Stage~2 training \\
    \midrule
    GRU encoder $\phi_{\text{gru}}$ and cross-asset encoder $\mathcal{E}$ & \checkmark & fixed \\
    Codebook $\mathcal{Z}$ (VQ embeddings) & \checkmark & fixed \\
    Decoder $\mathcal{G}$ and predictor $\psi_{\text{pred}}$ (aux. tasks) & \checkmark & fixed \\
    Temporal Transformer $\mathcal{T}$ & -- & \checkmark \\
    MoE (gating + experts) and loading head $\psi_{\text{load}}$ & -- & \checkmark \\
    \bottomrule
    \end{tabular}
\end{adjustbox}
\end{table}

Freezing the spatial representation after Stage~1 prevents codebook collapse and avoids co-adaptation between discrete structure learning and temporal dynamics, enforcing a clear information bottleneck between the two stages.

\subsection{Stage~1 (Spatial Learning): Quantizer, Decoder, and Auxiliary Prediction}
\label{app:stage1_impl}

This section describes the implementation details of Stage~1, which learns discrete cross-sectional representations and associated latent factor values.

\paragraph{Input normalization (RevIN).}
Before encoding, each stock’s input sequence $x_i\in\mathbb{R}^{T\times C}$ is normalized using reversible instance normalization (RevIN)~\cite{kim2021reversible}.
RevIN normalizes each time series independently using per-sample statistics and applies the corresponding inverse transform to the reconstructed output.
This mitigates non-stationarity across stocks and time without leaking future information.

\paragraph{Cross-asset encoding.}
At each date $t$, the GRU encoder $\phi_{\text{gru}}$ maps temporal features $x_i\in\mathbb{R}^{T\times C}$ to a fixed-length embedding $h_i\in\mathbb{R}^{d_s}$ (Eq.~\eqref{eq:gru_enc}). Stacking all stocks yields the cross-sectional matrix $H\in\mathbb{R}^{N_t\times d_s}$. This matrix is processed by the cross-asset Transformer $\mathcal{E}$, implemented as a stack of encoder blocks (Eq.~\eqref{eq:encoder_block}), to produce context-aware embeddings $z_i\in\mathbb{R}^{d_s}$ (Eq.~\eqref{eq:cross_asset}).
Self-attention enables each stock representation to incorporate information from related stocks in the same cross section.

\paragraph{Vector quantization.}
Each continuous embedding $z_i\in\mathbb{R}^{d_s}$ is discretized by assigning it to the nearest codeword $\bm{c}_k$ in the codebook $\mathcal{Z}=\{\bm{c}_k\}_{k=1}^{K}$:
\begin{align}
k_i &= \arg\min_{k\in\{1,\ldots,K\}} \|z_i-\bm{c}_k\|_2^2,
\quad
z_{q,i}=\bm{c}_{k_i}\in\mathbb{R}^{d_s}.
\end{align}
Training follows Eq.~\eqref{eq:vq_loss} with a commitment weight $\lambda_{\text{commit}}{=}0.25$ and the straight-through estimator.
To encourage balanced code usage, an exponential moving average (EMA) of code assignments is maintained, and underutilized codes are periodically re-initialized via probability-based anchoring. The codebook contrastive loss (Eq.~\eqref{eq:contra}) is applied with a temperature $\tau=0.07$, explicitly separating code prototypes in representation space.

\paragraph{FiLM-conditioned reconstruction decoder $\mathcal{G}$.}
The reconstruction decoder enforces that discrete codes preserve information necessary to recover the original input. Given a quantized code $z_{q,i}$ and prior factors $f_p$, the decoder reconstructs $\hat{x}_i=\mathcal{G}(z_{q,i},f_p)$ (Eq.~\eqref{eq:recon}) using a FiLM-conditioned multi-resolution 1D upsampling architecture.

First, $z_{q,i}\in\mathbb{R}^{d_s}$ is projected into a low-resolution hidden sequence:
\begin{equation}
\begin{split}
X^{(0)}_i &= \mathrm{reshape}\big(\mathrm{GELU}(W_{\mathrm{in}} z_{q,i})\big), \\
\text{where} \quad & X^{(0)}_i \in \mathbb{R}^{H\times T_0}, \quad
W_{\mathrm{in}}\in\mathbb{R}^{(H T_0) \times d_s },
\end{split}
\end{equation} where $H{=}128$ and $T_0{=}5$. We then apply $K_u=\log_2(T/T_0)$ upsampling blocks (with $T{=}20$, so $K_u{=}2$). Each block performs channel expansion, PixelShuffle upsampling, FiLM conditioning on $f_p$, and a transposed-convolution skip connection. FiLM generates per-channel affine parameters $(\gamma,\beta)\in\mathbb{R}^{H}$ and applies
$
X \leftarrow X\odot(1+\gamma)+\beta
$
in residual form.
A final $1{\times}1$ convolution projects the output to $C{=}158$ channels, yielding $\hat{x}_i\in\mathbb{R}^{T\times C}$.

\paragraph{Multi-horizon auxiliary prediction $\psi_{\text{pred}}$.}
To encourage predictive structure in the learned codes, Stage~1 includes an auxiliary multi-horizon return prediction task.
The predictor $\psi_{\text{pred}}(z_{q,i},f_p)$ (Eq.~\eqref{eq:pred}) is implemented as a GRU-based autoregressive decoder.
An MLP maps the concatenated vector $[z_{q,i};f_p]$ to the initial GRU hidden state.
A learnable start token is used to autoregressively generate $N_h$ future returns.
This auxiliary objective is weighted by $\lambda_{\text{pred}}{=}10^{-4}$ and serves as a weak supervision signal that aligns discrete codes with future return dynamics.

\subsection{Stage~2 (Temporal Learning): Temporal Encoding and Structure-Conditioned MoE}
\label{app:stage2_impl}

Stage~2 focuses on modeling time-varying dynamics and generating dynamic factor loadings conditioned on both temporal information and learned cross-sectional structure.
Unlike Stage~1, which is frozen after training, all parameters in Stage~2 are optimized end-to-end.

\paragraph{Temporal encoder $\mathcal{T}$.}
Stage~2 consumes the same input window $x_i\in\mathbb{R}^{T\times C}$ as Stage~1, together with the discrete code $z_{q,i}\in\mathbb{R}^{d_s}$ produced by the fixed codebook.
Following Eq.~\eqref{eq:temp_enc}, cross-sectional structure is injected into temporal modeling by prepending a \emph{structure token} derived from $z_{q,i}$.
When $d_s \neq d_t$, the code is first projected to $\mathbb{R}^{d_t}$ via a linear layer.

The resulting token sequence is processed by a Transformer encoder $\mathcal{T}$ equipped with Rotary Position Embeddings (RoPE; Appendix~\ref{app:rope}), which allows attention scores to depend on relative temporal positions.
The temporal stream is first linearly mapped to $\mathbb{R}^{d_t}$ and then passed through $\mathcal{T}$ (1 encoder layer; FFN dimension 128; dropout 0.1).
We use 2 attention heads for CSI~300 and 4 heads for S\&P~500.
The output corresponding to the structure token is taken as a compact temporal summary
$h_{\mathrm{temp},i}\in\mathbb{R}^{d_t}$, which aggregates both temporal dynamics and cross-sectional context.

\paragraph{Structure-conditioned MoE and loading head $\psi_{\text{load}}$.}
Dynamic factor loadings are generated using a structure-conditioned Mixture-of-Experts (MoE).
The MoE decouples \emph{expert selection}, which depends on discrete structural codes, from \emph{expert computation}, which depends on temporal information.
This design allows different experts to specialize in distinct market regimes associated with different codes.

\paragraph{Fusion to form expert inputs.}
The temporal summary $h_{\mathrm{temp},i}$ and discrete code $z_{q,i}$ are first normalized:
$\tilde{h}_i=\mathrm{LN}(h_{\mathrm{temp},i})\in\mathbb{R}^{d_t}$ and
$\tilde{z}_i=\mathrm{LN}(z_{q,i})\in\mathbb{R}^{d_s}$.
These are concatenated and projected through a small residual MLP to form the expert input:
\begin{align}
u_i = \phi_{\mathrm{proj}}([\tilde{h}_i;\tilde{z}_i]) \in \mathbb{R}^{d_t}.
\end{align}
The projection module $\phi_{\mathrm{proj}}$ consists of a residual feed-forward block with GEGLU-style gating and dropout (0.1), which improves expressiveness while maintaining training stability.

\paragraph{Code-conditioned sparse routing.}
Expert routing depends solely on the discrete code, enforcing a clear separation between structural conditioning and temporal computation. 
The gate computes logits from $\tilde{z}_i$ and selects the top-$k$ experts (stochastic via Eqs.~(\ref{eq:gate_mu_sigma})--(\ref{eq:gate_reparam}) at training, deterministic $\ell_i = \mu_i$ at inference):
\begin{align}
\ell_i &= \phi_{\mathrm{gate}}(\tilde{z}_i)\in\mathbb{R}^{M_e}, \qquad
\mathcal{K}_i=\mathrm{Top}_k(\ell_i), \nonumber\\
G_{i,j}
&=
\begin{cases}
\displaystyle\frac{\exp(\ell_{i,j})}{\sum_{\ell\in\mathcal{K}_i}\exp(\ell_{i,\ell})} & \text{if } j\in\mathcal{K}_i,\\[6pt]
0 & \text{otherwise.}
\end{cases}
\label{eq:app_topk_gate}
\end{align}
This yields a sparse probability vector $G_i$ with $\|G_i\|_0=k$ and $\sum_j G_{i,j}=1$.
Each expert $\xi_j$ is a lightweight MLP mapping $u_i\in\mathbb{R}^{d_t}$ to $\mathbb{R}^{d_{\mathrm{moe}}}$, with hidden size $d_{\mathrm{moe}}{=}64$ and dropout 0.1.
Expert outputs are aggregated as
\begin{align}
m_i = \sum_{j=1}^{M_e} G_{i,j}\,\xi_j(u_i)\in\mathbb{R}^{d_{\mathrm{moe}}}.
\label{eq:app_moe_agg}
\end{align}
We use $M_e{=}2$ experts with top-1 routing for CSI~300, and $M_e{=}8$ experts with top-4 routing for S\&P~500, reflecting differences in market complexity and signal heterogeneity.

\paragraph{Dynamic loadings via base--modulation decomposition.}
Rather than predicting $(\alpha_i,\beta_{p,i},\beta_{l,i})$ directly, the loading head decomposes generation into a temporal \emph{base} and a code-conditioned \emph{modulation}.
Base loadings are derived from temporal information:
\begin{align}
\beta^{\mathrm{base}}_{p,i} = W^{\mathrm{base}}_{p} h_{\mathrm{temp},i}\in\mathbb{R}^{P}, \quad
\beta^{\mathrm{base}}_{l,i} = W^{\mathrm{base}}_{l} h_{\mathrm{temp},i}\in\mathbb{R}^{d_s}.
\end{align}
The MoE output $m_i$ predicts FiLM-style modulation parameters:
\begin{align}
(\gamma_{p,i},\delta_{p,i}) &= \phi_{p}(m_i)\in\mathbb{R}^{2P}, \nonumber\\
(\gamma_{l,i},\delta_{l,i}) &= \phi_{l}(m_i)\in\mathbb{R}^{2d_s}.
\end{align}
Final dynamic loadings are produced by residual affine modulation:
\begin{align}
\beta_{p,i} = \gamma_{p,i}\odot \beta^{\mathrm{base}}_{p,i} + \delta_{p,i}, \quad
\beta_{l,i} = \gamma_{l,i}\odot \beta^{\mathrm{base}}_{l,i} + \delta_{l,i}.
\label{eq:hyperfusion_beta}
\end{align}
The intercept is predicted as $\alpha_i = w_\alpha^\top m_i$.
This decomposition improves stability and interpretability by separating time-varying effects from structure-dependent adjustments.

\paragraph{Regularization and training objective.}
Stage~2 is trained using the temporal objective in Eq.~\eqref{eq:temporal_obj}.
In addition to the standard MoE load-balancing auxiliary loss, an $\ell_2$ penalty is applied to the loadings,
$\|\beta_{p,i}\|_2+\|\beta_{l,i}\|_2$, to discourage excessive exposure magnitudes.
Final return prediction follows Eq.~\eqref{eq:final_pred}.


\subsection{Optimization and Reproducibility}
\label{app:optim_repro}

PRISM-VQ is implemented in PyTorch and optimized using AdamW with learning rate $10^{-4}$. Gradient clipping with threshold 1.0 is applied, and early stopping is used based on validation performance. All reported results are averaged over five random seeds $\{0,1,2,3,4\}$ to ensure statistical robustness.


\section{Expert Prior Factors}
\label{app:prior_factors}

PRISM-VQ incorporates a set of expert-designed financial factors as domain-informed priors. Specifically, we use 13 factor return series from the JKP Global Factor Library\footnote{\url{https://jkpfactors.com/}}, a widely used and publicly documented collection of global equity factors \cite{jensen2023there}. The selected factors span multiple dimensions of firm characteristics and return anomalies:
\begin{itemize}
    \item \texttt{accruals}: Measures the extent to which earnings are driven by accounting accruals rather than cash flows; high accruals are often associated with lower future returns.
    \item \texttt{debt\_issuance}: Captures changes in corporate leverage through net debt issuance; firms issuing more debt tend to underperform subsequently.
    \item \texttt{investment}: Reflects asset growth or capital investment intensity; high investment rates are empirically linked to lower expected returns.
    \item \texttt{low\_leverage}: Favors firms with lower financial leverage, which are typically less exposed to distress risk.
    \item \texttt{low\_risk}: Proxies for low-volatility or low-beta characteristics; low-risk stocks have been shown to earn anomalously high risk-adjusted returns.
    \item \texttt{momentum}: Measures past intermediate-horizon returns; stocks with strong recent performance tend to continue outperforming in the short to medium term.
    \item \texttt{profit\_growth}: Captures growth in operating profitability; firms with improving profitability often experience positive revaluation.
    \item \texttt{profitability}: Measures the level of operating profits relative to assets or equity; highly profitable firms tend to earn higher average returns.
    \item \texttt{quality}: Aggregates multiple firm characteristics such as profitability, earnings stability, and conservative investment behavior.
    \item \texttt{seasonality}: Exploits recurring calendar-based return patterns, such as month-of-the-year effects.
    \item \texttt{short\_term\_reversal}: Captures mean-reversion in very recent returns; stocks that underperform in the short term often rebound.
    \item \texttt{size}: Represents firm size, typically measured by market capitalization; smaller firms historically earn higher average returns.
    \item \texttt{value}: Measures valuation ratios such as book-to-market; cheaper stocks tend to outperform growth stocks over long horizons.
\end{itemize}

Each factor is treated as a \emph{market-wide factor return series}, rather than as a stock-specific characteristic. These factors serve as economically interpretable anchors that encode well-established return drivers from asset pricing theory.
Throughout construction, we strictly enforce a no–look-ahead constraint: all factor values used at decision time $t$ are computed exclusively from information available up to $t-1$.

\paragraph{Rolling cumulative return construction (excluding current day).}
Let $r_{j,\tau}$ denote the daily return of factor $j$ on trading date $\tau$.
To align factor inputs with the model lookback window $T{=}20$ and ensure temporal consistency with stock-level features, we construct prior factor inputs as rolling cumulative returns over the previous $T$ trading days, excluding the current day.

Concretely, factor $j$ at decision time $t$ is defined as
\begin{align}
f_{p,j}(t)
&= \exp\!\left(\sum_{\tau=t-T}^{t-1} \log\!\left(1+r_{j,\tau}\right)\right) - 1.
\label{eq:prior_20d_cum}
\end{align}
This formulation is equivalent to the multiplicative return
$\prod_{\tau=t-T}^{t-1} (1+r_{j,\tau}) - 1$,
implemented in log space for numerical stability.
By shifting the window to end at $t-1$, the construction explicitly excludes contemporaneous factor information, preventing any form of look-ahead bias.

\paragraph{Normalization and usage.}
All factor series are standardized using statistics computed on the training period only. The resulting vector $f_p(t)\in\mathbb{R}^{P}$ is provided as an input to both Stage~1 and Stage~2. In Stage~1, prior factors condition reconstruction and auxiliary prediction, encouraging discrete codes to remain compatible with known economic structure. In Stage~2, they enter the factor pricing equation directly via $\beta_{p,i}^\top f_p(t)$, preserving a transparent link between expert priors and predicted returns.

This design allows PRISM-VQ to leverage established financial knowledge for stability and interpretability while still learning complementary structure-driven signals through vector-quantized latent factors.

\section{Evaluation Metrics}
\label{app:metrics}

This appendix details the evaluation metrics used in Section~\ref{sec:experiments}.
Model performance is assessed from two complementary perspectives:
(i) cross-sectional ranking quality and
(ii) portfolio-level utility under a fixed trading protocol.
All expectations and standard deviations are computed over the test period.
The number of tradable stocks $N_t$ may vary over time due to index reconstitution.

\subsection{Ranking Performance Metrics}

We evaluate cross-sectional ranking accuracy using RankIC and its information ratio, RankICIR.
The model output $\hat{y}_{i,t}$ is interpreted as a \emph{ranking score} rather than a calibrated return forecast.
Accordingly, metrics focus on the relative ordering of stocks within each date.

\paragraph{RankIC.}
RankIC is defined as the Spearman rank correlation between predicted scores and realized returns in the cross section.
At each date $t$, we compute RankIC by applying the Pearson correlation to within-date ranks:
\begin{align}
    \text{RankIC}_t
    &= \frac{(r_{\hat{y}_t} - \bar{r}_{\hat{y}_t})^\top (r_{y_t} - \bar{r}_{y_t})}
    {N_t \cdot \sigma_{\hat{y}_t} \cdot \sigma_{y_t}},
\end{align}
where $r_{\hat{y}_t}, r_{y_t} \in \mathbb{R}^{N_t}$ denote the within-date ranks of predicted scores and realized returns, respectively (rank 1 corresponds to the highest value).
Here, $\bar{r}$ and $\sigma$ denote the within-date mean and standard deviation of ranks.
This formulation ensures that RankIC is invariant to monotonic transformations of $\hat{y}_{i,t}$ and is robust to scale differences across dates.

The reported RankIC is the time average over the test period:
\begin{align}
    \text{RankIC}
    &= \frac{1}{T_{\text{test}}} \sum_{t=1}^{T_{\text{test}}} \text{RankIC}_t.
\end{align}

\paragraph{RankICIR.}
To measure the stability and consistency of ranking performance over time, we report RankICIR, defined as
\begin{align}
    \text{RankICIR}
    &= \frac{\mathbb{E}[\text{RankIC}_t]}{\mathrm{std}(\text{RankIC}_t)}.
\end{align}
RankICIR penalizes models with volatile daily ranking performance, even if their average RankIC is high.
This metric is particularly relevant in financial applications, where consistent signal quality is crucial for portfolio construction and risk management.

Overall, RankIC captures the average cross-sectional predictive power, while RankICIR reflects its temporal reliability.

\subsection{Portfolio Performance Metrics}

Portfolio performance is evaluated using the same Top$K$--Drop$N$ rebalancing protocol described in Section~\ref{sec:experiments}.
At each date $t$, let $\mathcal{P}_t$ denote the set of held stocks after ranking and filtering, and let $w_{i,t}$ be the portfolio weight assigned to stock $i\in\mathcal{P}_t$.
Unless otherwise specified, portfolios are equal-weighted, i.e., $w_{i,t}=1/|\mathcal{P}_t|$.

\paragraph{Daily portfolio return.}
Given realized stock returns $y_{i,t}$, the portfolio log return is computed as
\begin{equation}
g_{p,t}
= \log\!\left(1 + \sum_{i \in \mathcal{P}_t} w_{i,t}\, y_{i,t}\right),
\end{equation}
which ensures time additivity and numerical stability when aggregating returns across periods.

\paragraph{Annualized return (AR).}
Annualized return is computed from the mean daily log return assuming 252 trading days per year:
\begin{equation}
\text{AR}
= \exp\!\big(252 \cdot \mathbb{E}[g_{p,t}]\big) - 1.
\end{equation}

\paragraph{Maximum drawdown (MDD).}
Let $W_t = \exp\!\left(\sum_{\tau=1}^{t} g_{p,\tau}\right)$ denote the cumulative wealth process. Maximum drawdown measures the largest peak-to-trough loss over the evaluation horizon:
\begin{equation}
\text{MDD}
= \max_{t}\!\left(1 - \frac{W_t}{\max_{s \le t} W_s}\right).
\end{equation}

MDD captures downside risk and is particularly relevant for evaluating robustness under adverse market conditions.

\paragraph{Sharpe ratio (SR).}
Risk-adjusted performance is measured using the Sharpe ratio:
\begin{equation}
\text{SR}
= \frac{\sqrt{252} \cdot \mathbb{E}[g_{p,t}]}{\mathrm{std}(g_{p,t})},
\end{equation} where the denominator is the standard deviation of daily portfolio log returns. The risk-free rate is set to zero, following common practice in comparative ranking-based studies.

\paragraph{Transaction costs.}
Transaction costs are applied at each rebalance according to the protocol specified in Section~\ref{sec:experiments}, and are fully incorporated into $g_{p,t}$ before computing all reported portfolio metrics.

\section{Portfolio Construction}
\label{app:topk}

Portfolio performance is evaluated using Qlib’s \texttt{TopkDropoutStrategy}\footnote{\url{https://qlib.readthedocs.io/en/latest/component/strategy.html}}, also referred to as the Top$K$--Drop$N$ protocol.
This strategy converts cross-sectional ranking scores into a realistic, turnover-controlled trading portfolio.

\paragraph{Ranking-based candidate selection.}
Let $\mathcal{U}_t$ denote the tradable universe at trading day $t$, and let $\hat{y}_{i,t}$ be the predicted ranking score for stock $i$.
We first identify the top-ranked candidate set
\begin{equation}
\mathcal{C}_t
= \operatorname{TopK}\!\left(\{\hat{y}_{i,t}\}_{i \in \mathcal{U}_t},\, K_{\text{port}}\right),
\end{equation} which contains the $K_{\text{port}}$ stocks with the highest predicted scores on day $t$.

\paragraph{Turnover-constrained rebalancing.}
Let $\mathcal{P}_{t-1}$ denote the portfolio holdings from the previous trading day.
The Top$K$--Drop$N$ strategy updates the portfolio in two steps:
(i) it sells up to $N_{\text{drop}}$ currently held stocks with the lowest scores, and (ii) it replaces them with the same number of higher-ranked stocks from $\mathcal{C}_t$ that are not already held.
This procedure enforces an explicit turnover constraint by limiting the number of daily replacements to at most $N_{\text{drop}}$.

\paragraph{Portfolio weights.}
After rebalancing, the portfolio size is restored to $K_{\text{port}}$, and equal weights are assigned:
$w_{i,t} = 1 / |\mathcal{P}_t|$ for all $i \in \mathcal{P}_t$.
Unless otherwise stated, all experiments use equal-weighted portfolios.

\paragraph{Algorithmic description.}
Algorithm~\ref{alg:portfolio} provides a precise procedural description of the Top$K$--Drop$N$ strategy.

\begin{algorithm}[t]
\caption{\texorpdfstring{Top$K$--Drop$N$}{TopK--DropN}}
\label{alg:portfolio}
\small
\begin{algorithmic}[1]
\Require Scores $\hat{y}_t$, universe $\mathcal{U}_t$, previous holdings $\mathcal{P}_{t-1}$, $K_{\text{port}}$, $N_{\text{drop}}$
\Ensure Holdings $\mathcal{P}_t$ and equal weights $w_t$
\State $\mathcal{C}_t \gets \text{Top-}K_{\text{port}}\text{ stocks in }\mathcal{U}_t\text{ by }\hat{y}_{i,t}$
\State $\mathcal{H}_t \gets \mathcal{P}_{t-1} \cap \mathcal{U}_t$ \Comment{remove names leaving the universe}
\State $\mathcal{S}_t \gets \text{bottom-}N_{\text{drop}}\text{ stocks in }\mathcal{H}_t\text{ by }\hat{y}_{i,t}$ \Comment{sell set}
\State $\mathcal{H}_t \gets \mathcal{H}_t \setminus \mathcal{S}_t$
\State $\mathcal{B}_t \gets \text{top-}|\mathcal{S}_t|\text{ stocks in }(\mathcal{C}_t \setminus \mathcal{H}_t)\text{ by }\hat{y}_{i,t}$ \Comment{buy set}
\State $\mathcal{P}_t \gets \mathcal{H}_t \cup \mathcal{B}_t$
\State $w_{i,t} \gets 1/|\mathcal{P}_t|$ for all $i \in \mathcal{P}_t$
\end{algorithmic}
\end{algorithm}

\section{Transaction-Cost Robustness}
\label{app:tcrobust}

\begin{figure*}[t]
  \centering
  \begin{minipage}[t]{0.4\linewidth}
    \centering
    \includegraphics[width=\linewidth]{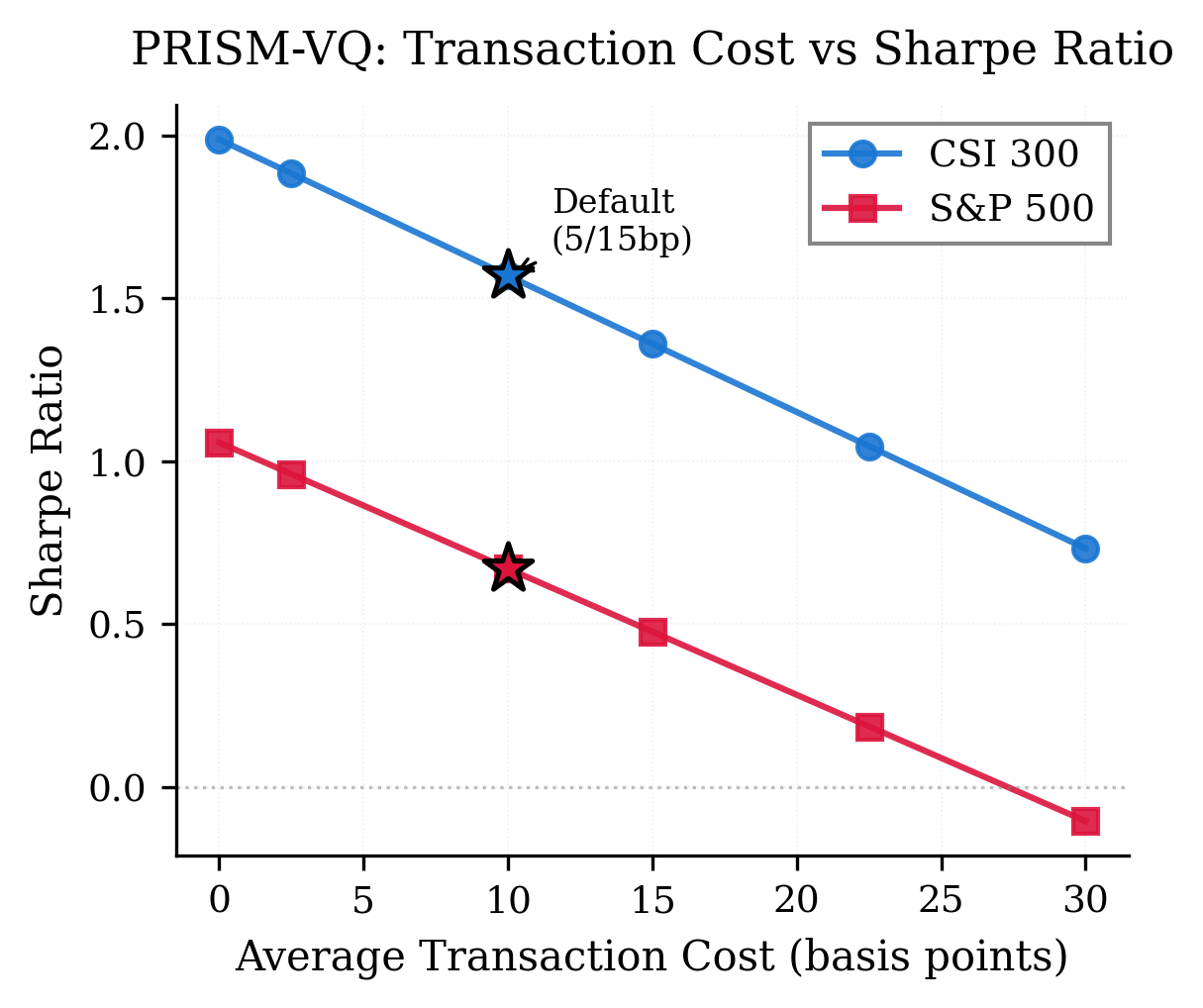}
    \vspace{2pt}
    {\footnotesize \textbf{(a) Sharpe ratio vs.\ turnover}}
  \end{minipage}\hfill
  \begin{minipage}[t]{0.4\linewidth}
    \centering
    \includegraphics[width=\linewidth]{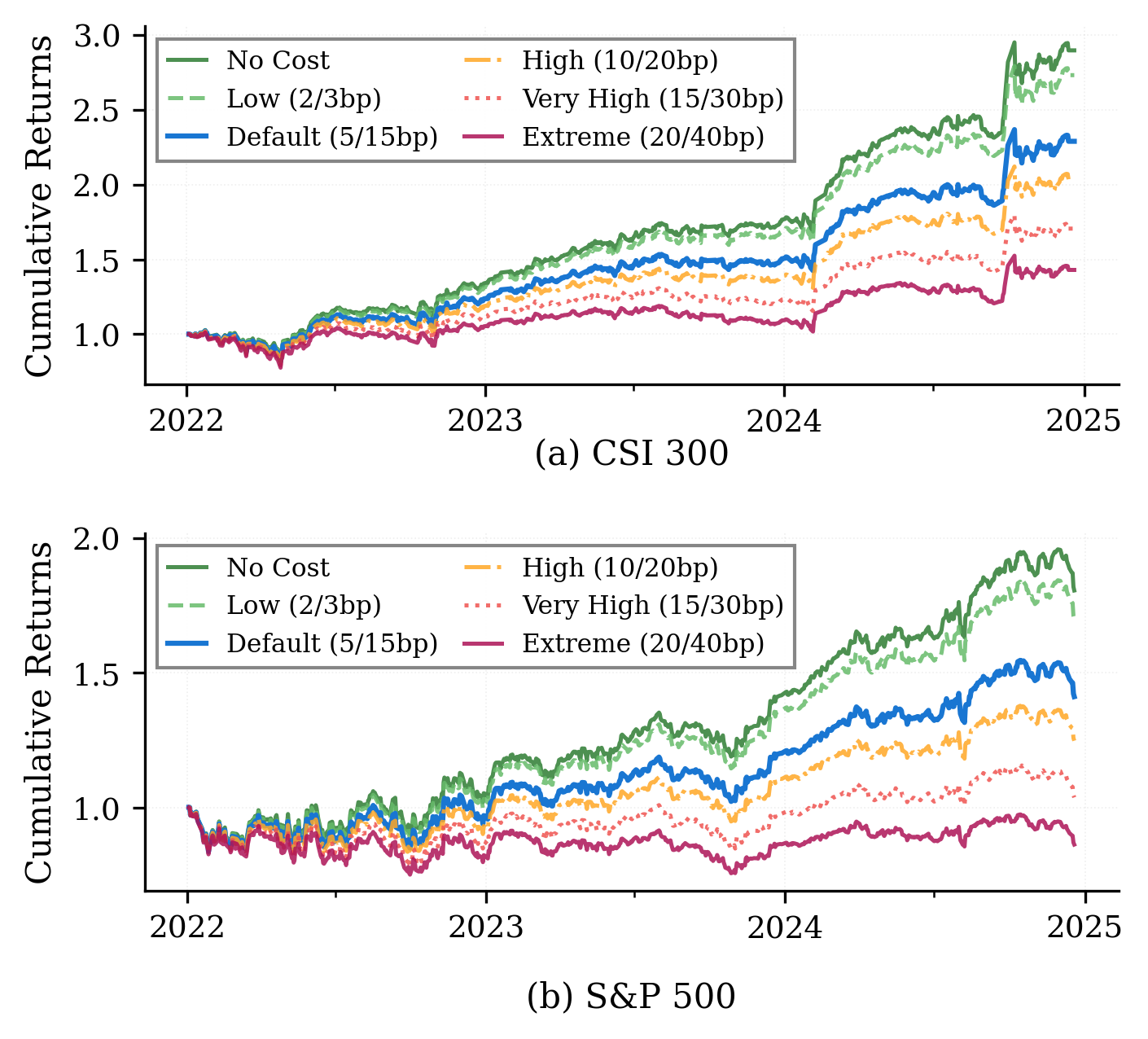}
    \vspace{2pt}
    {\footnotesize \textbf{(b) Cumulative return sensitivity}}
  \end{minipage}
  \vspace{2pt}
  \caption{Transaction-cost robustness of PRISM-VQ under identical prediction signals and portfolio constraints as in the main experiments.}
  \label{fig:tcost_robustness_panel}
  \vspace{-2mm}
\end{figure*}

To evaluate economic robustness, transaction costs are varied while holding all other experimental settings fixed.
Specifically, the stock universe, prediction signals, and portfolio construction protocol (Top$K$--Drop$N$ with fixed $K$ and $N$) are identical to those used in the main experiments.
Only proportional transaction costs applied at portfolio rebalancing are swept.

\paragraph{Transaction-cost regimes.}
Six cost scenarios are considered: \textit{No Cost}, \textit{Low} (2/3 bp), \textit{Default} (5/15 bp), \textit{High} (10/20 bp), \textit{Very High} (15/30 bp), and \textit{Extreme} (20/40 bp),
where entry and exit costs are charged asymmetrically (buy/sell).
All backtests follow the same daily rebalancing schedule and enforce the same turnover constraint implied by the Top$K$--Drop$N$ strategy.

\paragraph{Results and interpretation.}
Figure~\ref{fig:tcost_robustness_panel}(a) reports the Sharpe--turnover frontier across cost regimes, illustrating how risk-adjusted performance degrades as trading frictions increase.
Figure~\ref{fig:tcost_robustness_panel}(b) shows cumulative return trajectories under the same prediction signals while varying transaction costs, enabling a direct sensitivity analysis without altering model outputs.

Across both views, PRISM-VQ maintains competitive risk-adjusted performance under realistic cost levels and exhibits a smooth, monotonic degradation as costs rise.
This behavior indicates that the reported gains are not driven by excessive turnover or fragile short-horizon trading, but instead reflect robust cross-sectional signal quality.


\section{Sensitivity to TopK--DropN Rebalancing Rate}
\label{app:nsweep}

Section~\ref{sec:experiments} fixes $N_{\mathrm{drop}}{=}5$. We here vary $N_{\mathrm{drop}}$ at $K_{\mathrm{port}}{=}30$ with all other settings unchanged. Table~\ref{tab:nsweep} shows that AR and Sharpe rise with $N_{\mathrm{drop}}$, but mean daily turnover 
grows $\sim$14$\times$ from $N{=}1$ to $N{=}15$. At round-trip costs of 20--50~bps, the marginal gains for $N \geq 10$ are 
largely offset by trading frictions. The default $N{=}5$ thus offers a favorable net-of-cost balance.

\begin{table}[t]
\centering
\scriptsize
\setlength{\tabcolsep}{4pt}
\caption{Portfolio performance under varying $N_{\mathrm{drop}}$ at $K_{\mathrm{port}}{=}30$ (5-seed average). \textit{Cum.}: cumulative return over the test period; \textit{Turn.}: mean daily turnover. The default $N{=}5$ is shown in bold.}
\label{tab:nsweep}
\begin{tabular}{cccccccc}
\toprule
Market & $N$ & $N/K$ & AR$\uparrow$ & SR$\uparrow$ & MDD$\downarrow$ & Cum.$\uparrow$ & Turn. \\
\midrule
\multirow{6}{*}{\textbf{S\&P 500}} 
 & 1  & 0.0333 & 0.0172 & 0.0798 & 0.2974 & 0.0514 & 0.0662 \\
 & 3  & 0.1000 & 0.1135 & 0.5268 & 0.1605 & 0.3383 & 0.2005 \\
 & 5  & 0.1667 & 0.1442 & 0.6701 & 0.1616 & 0.4299 & 0.3310 \\
 & 7  & 0.2333 & 0.1772 & 0.8354 & 0.1656 & 0.5280 & 0.4592 \\
 & 10 & 0.3333 & 0.1913 & 0.9149 & 0.1600 & 0.5700 & 0.6511 \\
 & 15 & 0.5000 & 0.2044 & 1.0016 & 0.1487 & 0.6091 & 0.9392 \\
\midrule
\multirow{6}{*}{\textbf{CSI 300}} 
 & 1  & 0.0333 & 0.1931 & 1.1479 & 0.1886 & 0.5549 & 0.0658 \\
 & 3  & 0.1000 & 0.2636 & 1.4411 & 0.1905 & 0.7574 & 0.1973 \\
 & 5  & 0.1667 & 0.3077 & 1.5694 & 0.1924 & 0.8841 & 0.3259 \\
 & 7  & 0.2333 & 0.3035 & 1.5140 & 0.1879 & 0.8720 & 0.4443 \\
 & 10 & 0.3333 & 0.3179 & 1.5635 & 0.1833 & 0.9133 & 0.5662 \\
 & 15 & 0.5000 & 0.3006 & 1.4864 & 0.1940 & 0.8637 & 0.6183 \\
\bottomrule
\end{tabular}
\end{table}

\section{Stock Universe for t-SNE}
\label{app:stocks}

\begin{table}[t]
\centering
\caption{Representative stocks used in the t-SNE visualization (41 stocks across 10 sectors).}
\label{tab:tsne_stocks}
\scriptsize
\setlength{\tabcolsep}{3pt}
\renewcommand{\arraystretch}{1.05}
\begin{tabular}{p{0.3\linewidth} p{0.63\linewidth}}
\toprule
Sector & Tickers \\
\midrule
Technology &
AAPL, MSFT, NVDA, GOOGL, META \\
Financials &
JPM, BAC, WFC, GS, MS \\
Healthcare &
JNJ, UNH, LLY, ABBV, MRK \\
Consumer &
AMZN, COST, HD, KO, NKE, PG, TSLA, WMT \\
Industrials &
BA, CAT, UPS, HON \\
Real Estate &
AMT, PLD, CCI \\
Energy &
COP, CVX, XOM \\
Materials &
LIN, APD, SHW \\
Communication Services &
DIS, NFLX \\
Utilities &
DUK, NEE, SO \\
\bottomrule
\end{tabular}
\end{table}

Learned discrete representations are visualized using t-SNE on a random subsample of observations from the test period.
To improve interpretability and ensure consistency across visualization runs, a fixed set of 41 representative large-cap stocks spanning 10 broad GICS-style sectors is explicitly included in each plot.

These stocks act solely as visual anchors: they do not influence model training, inference, or evaluation.
Their inclusion facilitates qualitative inspection of whether the learned discrete codes align with, or instead cut across, conventional sector classifications.
Table~\ref{tab:tsne_stocks} lists the full anchor set used in Figure~\ref{fig:tsne}.

\section{Code Transition Dynamics}
\label{app:transition}

\begin{figure*}[htbp]
    \centering
    \includegraphics[width=0.7\linewidth]{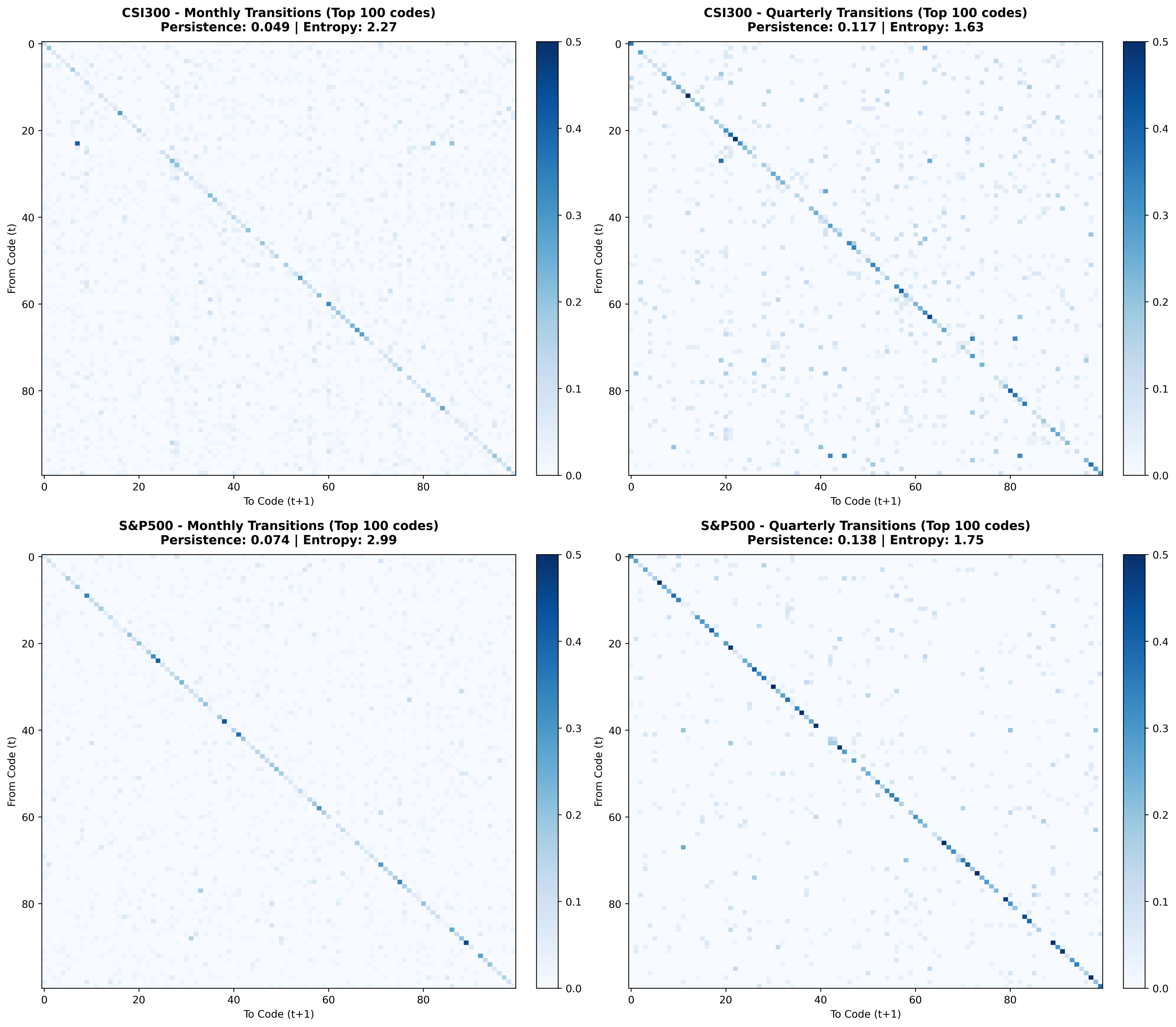}
    \caption{Code transition probability matrices over monthly (left) and quarterly (right) horizons for the top 100 most active codes in CSI~300 (top) and S\&P~500 (bottom). Diagonal entries indicate persistence, while entropy summarizes transition dispersion.}
    \label{fig:transition}
\end{figure*}

To assess the temporal stability of the learned discrete representations, code transition probabilities are analyzed over monthly and quarterly horizons.
Figure~\ref{fig:transition} reports empirical transition matrices for the top 100 most frequently assigned codes in each market.
Diagonal dominance reflects persistence: at the monthly horizon, the average diagonal mass is 4.9\% for CSI~300 and 7.4\% for S\&P~500, substantially exceeding the uniform baseline of 1\%.
Persistence further increases at the quarterly horizon (11.7\% for CSI~300 and 13.8\% for S\&P~500), indicating that the learned codes capture stable cross-sectional structure rather than transient noise.

The two markets exhibit distinct transition characteristics.
CSI~300 shows lower persistence but more concentrated transitions, with lower transition entropy (2.27), suggesting that stocks switch codes more frequently but tend to move among a limited set of successor states.
In contrast, S\&P~500 exhibits higher persistence together with more dispersed transitions (entropy 2.99), indicating that stocks remain in the same code for longer periods but explore a broader set of alternative states when transitions occur.
These differences are consistent with the market-specific structure-conditioned MoE configurations described in the main text, where the more efficient U.S. market benefits from greater expert diversity.

\end{document}